\documentclass[letterpaper]{article} % DO NOT CHANGE THIS
\usepackage{aaai24}  % DO NOT CHANGE THIS
\usepackage{times}  % DO NOT CHANGE THIS
\usepackage{helvet}  % DO NOT CHANGE THIS
\usepackage{courier}  % DO NOT CHANGE THIS
\usepackage[hyphens]{url}  % DO NOT CHANGE THIS
\usepackage{graphicx} % DO NOT CHANGE THIS
\usepackage{natbib}  % DO NOT CHANGE THIS AND DO NOT ADD ANY OPTIONS TO IT
\usepackage{caption} % DO NOT CHANGE THIS AND DO NOT ADD ANY OPTIONS TO IT
\usepackage{algorithm} 
\usepackage{algorithmic}  
\usepackage[algo2e]{algorithm2e} 
\usepackage{amsmath,amssymb}
\usepackage{mathtools}
\usepackage{enumitem}
\usepackage{amsfonts} 
\usepackage{caption}
\usepackage{subcaption}
\usepackage{booktabs}
\usepackage{multirow}
\usepackage{color}
\usepackage{newfloat}
\usepackage{listings}
\DeclareMathOperator{\EX}{\mathbb{E}}
\DeclareCaptionStyle{ruled}{labelfont=normalfont,labelsep=colon,strut=off} % DO NOT CHANGE THIS
\floatstyle{ruled}
\newfloat{listing}{tb}{lst}{}
\floatname{listing}{Listing}
\title{Personalized Path Recourse}
\author{
    Dat Hong\textsuperscript{\rm 1},
    Tong Wang\textsuperscript{\rm 2}
}
\affiliations{
    \textsuperscript{\rm 1}Yale University, School of Management\\
    \textsuperscript{\rm 2}Yale University, School of Management\\
    dat-hong@yale.edu, tong.wang.tw687@yale.edu
}

\usepackage{bibentry}
\begin{document}

\maketitle

\begin{abstract}
This paper introduces Personalized Path Recourse, a novel method that generates recourse paths for a path from an agent. The objective is to \textit{edit} a given path of actions to achieve desired goals (e.g., better outcomes compared to the agent’s original paths of actions) while ensuring the new path is personalized to the agent. Personalization refers to the extent to which the new path is similar to the original path and tailored to the agent’s observed behavior patterns from their policy function. We train a personalized recourse agent to generate such personalized paths, which are obtained using reward functions that consider the goal and personalization. The proposed method is applicable to both reinforcement learning and supervised learning settings for correcting or improving sequences of actions or sequences of data to achieve a pre-determined goal. The method is evaluated in various settings and demonstrates promising results.
\end{abstract}

\section{Introduction}

Imagine John, a 30-year-old accountant earning an annual income of \$50,000, who wishes he had pursued a different life path that could have led to a higher income of over \$100,000. This situation raises a fundamental question: in a multi-step decision-making setting, given a current path that is considered inferior, how can we generate alternative paths of actions that might have resulted in a better outcome? %In supervised learning, this problem is similar to generating counterfactual explanations or recourse. In reinforcement learning, we refer to this process as creating alternative paths of actions that could have led to a better reward or outcome, which we call "recourse."

There are many alternative paths that John could have taken to earn more than \$100,000, such as pursuing a computer science degree when he was in college or accepting a different job offer a couple of years ago, or, more radically, starting sports training when he was young to become a professional athlete. However, with the \emph{goal} of making more money, which path is the most suitable for John? The problem is not just to find the best path for achieving the goal, the problem is to find the best path \emph{for John} to achieve the goal. In other words, the new path needs to be \textbf{personalized} for John.

We define the meaning of personalization in the context of multi-step decision making.
 First, people generally prefer options that aren’t too different from their initial choices, as these familiar options tend to be more comfortable and less risky. For example, a driver is likely to prefer a path similar to their usual one, as they are familiar with the surroundings and traffic conditions. Similarly, in a coaching context, individuals are more receptive to feedback based on their past performance rather than being advised to adopt an entirely new strategy. This means when achieving the desired goal, we prefer smaller changes to John's original path. Thus the new path needs to bear as much similarity as possible to the original path.
 We call this \textit{path-level} personalization. 

 Another aspect of personalization we consider the behaviors or policies of individuals. Returning to John's career example, a new path should be tailored specifically to him, respecting his behavior patterns and preferences. For instance, if John enjoys mathematics (as reflected in his current job as an accountant) but lacks athletic talent, recommending a path that leads him to become a data scientist would be more appropriate than suggesting he become an athlete. This approach follows John’s intrinsic policy, which is shaped by his education history and behavior over time. We call this \emph{policy-level} personalization.

 Therefore, the path-level personalization encourages following the original path as much as possible and the policy-level personalization encourages following the original policy function as much as possible. 
We call the new path that could help John reach his \textit{goal} and is personalized for him at both path and policy level  \textbf{personalized recourse}.
% In practice, one can opt for any of those three factors or all, which we will illustrate those concepts in the experiment sections.

This personalized recourse is different from but related to counterfactual explanations or algorithmic recourse, which have predominantly focused on supervised learning \cite{verma2020counterfactual,pawlowski2020deep,tsirtsis2020decisions,goyal2019counterfactual}: how to change one or some of the features of an instance $\mathbf{X}$ to get a different outcome $Y$? %However, counterfactual explanations and algorithmic recourse are rarely applied to reinforcement learning \cite{gajcin2022counterfactual}.  
Out of the few relevant works we found for reinforcement learning, counterfactual reasoning is used for the purpose of   \emph{explaining} why an action is chosen for a given state $\mathbf{s}$ over the other actions, by generating a counterfactual state $\mathbf{s}^\prime$ that could lead to a different action \cite{olson2019counterfactual,madumal2020explainable,gajcin2023raccer}. 
However, these methods do not deviate from similar approaches in supervised learning because the explanations are on the state level and still focus on one decision (action) at a time \cite{gajcin2022counterfactual}. \citet{frost2022explaining} extends the explanation scope from states to paths, but still for the purpose of explanations: they create a set of informative paths of the agent's behavior under a test-time state distribution but these paths do not necessarily lead to a better outcome. 

None of the previously mentioned papers solve our problem, which focuses on ``recourse'' rather than ``explanation'', with the goal of creating an entire path of actions that can lead an agent to a desired goal (e.g., better outcome), while being tailored to the agent. In addition, none of the above work considers the policy level personalization and only studies counterfactuals at the path level.

In practical applications, policy-level personalization is crucial, which ensures that any recommended recourse respects a given \textit{policy}, describing how an individual would naturally act in different situations. For instance, an experienced driver might prefer the highway because it is faster, while a new driver might choose a local path because it feels safer. These concepts apply broadly: a gamer might want to find a new way to beat a game with a different play style, or a writer might wish to paraphrase a sentence to convey a different sentiment while maintaining a specific genre, such as scientific or fantasy. Nowadays, policies are readily available through pre-trained models or can be easily learned from large datasets. Some examples of this could be the pre-trained language models from a certain corpus, or dedicated system/database from streaming companies to query users’ choices when being given a set of recommendations. In the gaming industry, the policy can also be trained from the data collected from players over time.

In this paper, we introduce a method called Personalized Path Recourse (PPR), which works with sequence data modeled by Markov Decision Processes (MDP) \cite{puterman2014markov}. Our approach trains a new policy to generate recourse paths that pursue both path-level and policy-level personalization while leading to a better outcome. 
To train a personalized recourse agent, we design a reward function that takes into account three critical aspects: similarity to the original path, personalization to the agent’s behavior (as reflected in their policy), and the goal outcome.
Our framework is versatile, applicable to various settings, including reinforcement learning and supervised learning, where the challenge is to modify a sequence to achieve a certain goal while ensuring the generated sequence remains plausible. 
The method ensures scalability in large state and action spaces.

 %we have access to.  \textcolor{blue}{Dat: I believe we should no longer say $\mathcal{P}_A(.)$ is provided. Instead, we say this personalization policy is explicitly trained. The reason we train $\mathcal{P}_A(.)$ first is because the personalization information could come from different sources and in different format. Our goal is to build a universal framework that work for all scenarios.  $\mathcal{P}_A(.)$ is an appropriate representation of the personalization part because there are plenty of available methods to train $\mathcal{P}_A(.)$ from the data, or it is even usually already provided beforehand (language models, trained policies in reinforment learning,... ) }

% In a reinforcement learning setting,  agent $A$'s behavior is represented by her policy function $\mathcal{P}_A(\cdot)$, which is either directly provided or learned from past data.  

% Additionally, we implement an efficient exploration strategy to ensure scalability in large state and action spaces.

We evaluate PPR in different settings, comparing it against adapted and modified methods, as no existing methods directly address our problem. Our experiments demonstrate that PPR achieves more personalized paths while attaining the desired goal, effectively adapting to the unique styles of different agents.

\section{Related Work}
% % \subsection{Counterfactual Explanations}
% discuss paper 
% 1. [Counterfactual Explanations in Sequential Decision Making Under Uncertainty]
% 2. \cite{olson2019counterfactual}
% 3. Explaining Reinforcement Learning Policies through Counterfactual paths (not so related)
% 4. RACCER: Towards Reachable and Certain Counterfactual Explanations for Reinforcement Learning (discuss)
% This section reviews related work, starting from counterfactual explanations and recourse in a supervised learning setting, and then discusses related work in the setting of reinforcement learning.
\subsection{Algorithmic Recourse}
The problem PPR solves is closely related to the goal of
{algorithmic recourse} \cite{karimi2021algorithmic}, the process by which one can change an unfavorable outcome or decision made by an algorithm by altering certain input variables within their control. Algorithmic recourse can be considered as a special case of counterfactual explanations \cite{wachter2017counterfactual,dandl2020multi,thiagarajan2021designing}.
In this line of research, the input instance is represented by a fixed-length vector \cite{poyiadzi2020face,verma2022amortized,de2023synthesizing,wang2021learning,van2021conditional,sulem2022diverse,karlsson2020locally,hsieh2021dice4el,ates2021counterfactual}.
Some work considers the feasibility and costs of the changes made to the features \cite{ustun2019actionable,ross2021learning,joshi2019towards,upadhyay2021towards,harris2022bayesian,karimi2020survey}, and also personalization or user preference \cite{yetukuri2022actionable,de2023synthesizing}. 
However, these existing works have solely focused on \textit{supervised learning in tabular data settings}, where our method can work with both structured and unstructured data in a \textit{reinforcement learning} setting.
\subsection{Counterfactual Explanations for Reinforcement Learning}
The stream of work that is closely relevant to ours is counterfactual explanations, though they only focus on the \emph{path-level} similarity. 
The majority of the work on counterfactual explanations is for supervised learning environments.
Only a small number of research have been done for problems in reinforcement learning settings, and they are mostly \textit{focused on explanations for one action at a time}. Various methods \cite{olson2019counterfactual,gajcin2023raccer,madumal2020explainable} have been proposed to generate \emph{counterfactual states} that could lead to a different action when explaining an agent's current action for a given state.
% For instance, \citet{olson2019counterfactual} %, counterfactuals are generated at the state level using a generative deep-learning architecture. The authors 
% explains the decisions of RL agents in Atari games for each state by generating a counterfactual state to show what would need to change in a game image for the agent to choose a different action. Similarly, \citet{gajcin2023raccer} builds a tree to search for a counterfactual state that would obtain a different action, and \citet{madumal2020explainable} generates a counterfactual state by first learning a structural causal model during reinforcement learning and then generating explanations based on counterfactual analysis of the causal model. 
All these works provide counterfactual explanations for one state and action at a time. 
\citet{frost2022explaining} provides explanations at the path level, which trains an exploration policy to generate test time paths to explain how the agent behaves in unseen states following training. 
%Utilizing the exploration policy to generate explanation paths enables users to comprehend the agent's behavior when encountering new states. 
However, all these previously mentioned papers aim to \textit{explain rather than provide a complete solution, namely a new path, towards a better outcome}. %, while considering the similarity and personalization properties we care about.

% \textcolor{blue}{In addition to those lines of research, in reinforcement learning settings specifically, there is a set of works that propose frameworks to discover either more diverse or potentially better policies.\cite{racaniere2017imagination, zahavy2021discovering, andrychowicz2017hindsight,schaul2015universal}.
% However, while their ideas can help improve the agent's learning process, none of them could help solve finding an agent's personalized recourse path.}

\citet{tsirtsis2021counterfactual} is the most relevant to our task. The paper focuses on sequential decision-making under uncertainty and generates counterfactual explanations that specify an alternative sequence of actions differing in at most $k$ actions from the observed sequence that could have led to a better outcome. However, by employing dynamic programming and iterating through all the possible states and actions, the method only solves tasks with low-dimensional states and action spaces, as later shown in the experiments. Additionally, only comparing which actions are changed is not meaningful in many applications because the entire paths are already different after taking different actions, even if the following actions are the same. Finally, this method also does not consider personalization to an agent when looking for a counterfactual sequence of actions. This method will be later used as a baseline for some \textit{simple} tasks with low complexity.

\subsection{Alternative Path Planning}
In addition, there is prior work that find alternative policies for an agent by modifying its learning process \cite{racaniere2017imagination,zahavy2021discovering,andrychowicz2017hindsight}.
% For example, \cite{racaniere2017imagination} introduces Imagination-Augmented Agents to augment model-free agents by providing additional information from model-based planning.
%  \cite{zahavy2021discovering} proposed an algorithm to learn a more diverse set of policies by allowing the agent to learn a combination of intrinsic and extrinsic reward.
%  \cite{andrychowicz2017hindsight} improves the learned policy by relabeling failed experiences with actually achieved goals.
However, those works aim to discover either more diverse or potentially better policies. 
While their ideas can help improve the agent's training process, none of them could help solve finding an agent's personalized recourse path.

To the best of our knowledge, no prior work has been done to find a recourse path for an input path in the RL setting to achieve the desired outcome with personalized behavior. 

\section{Problem Formulation}
\label{method}
\subsection{Background and Notations}
\textcolor{black}{An MDP is a tuple M $\triangleq (\mathcal{S}, \mathcal{A}, P, R, \lambda$),} where  $\mathcal{S} = \{\mathbf{s}_1,...,\mathbf{s}_{|\mathcal{S}|}\}$ and $\mathcal{A} = \{a_1,...,a_{|\mathcal{A}|}\}$ are the set of  states and actions, respectively.
Here, $|\mathcal{S}|$, $|\mathcal{A}|$ are the cardinalities of set $\mathcal{S}$ and $\mathcal{A}$ respectively.
At time $t$, the agent at state $\mathbf{s}_t$ takes an action $a_t$ and moves to state $\mathbf{s}_{t+1}$ with a transition probability $P(\mathbf{s}_{t+1}|\mathbf{s}_t,a_t)$ and receives a reward, $r_t$, which is determined by a reward function $R(\mathbf{s}_t,a_t)$. 
% We also define a transition probability $P(\mathbf{s}_{t+1}|\mathbf{s}_t,a_t)$ measures the probability of transiting to state $\mathbf{s}_{t+1}$ from $\mathbf{s_t}$ after taking action $a_t$.
Behavior of the agent is defined by a policy function $\mathcal{P}_A(\mathbf{s}_t,a_t)$, which defines the probability that the agent $A$ will choose action $a_t$ given state $\mathbf{s}_t$.
% An episode is usually denoted by $\tau = [\mathbf{s_1},a_1,\mathbf{s_2},a_2,...,\mathbf{s_T}]$  ($T$ is the episode length), i.e., the agent was from state $\mathbf{s_1}$, took action $a_1$ then moved to state $\mathbf{s_2}$, and so on.
We denote $\tau = <\mathbf{s}_1,\mathbf{s}_2,...,\mathbf{s}_T>$ as a path, which is a sequence of states.
% In general, the goal of RL is to train the agent to learn an optimal, or nearly-optimal, policy $\pi(\mathbf{s},a)$  to help it maximize the reward from the environment.

% 
\paragraph{Deep Q-networks}
There are many algorithms for training a reinforcement learning agent.
In this paper, we employ deep Q-networks (DQN) to train a recourse policy. In a DQN algorithm, the agent explores the environment over multiple episodes, where each episode consists of a sequence of interactions with the environment from start to finish.
For each episode, at each time step $t$, we record the agent's current state $\mathbf{s}_t$, action $a_t$, the next state $\mathbf{s}_{t+1}$ and the reward $r_{t}$.
Those tuples $(\mathbf{s}_t,a_t,\mathbf{s}_{t+1},r_t)$ are stored in an experience replay $B$.
DQN trains a deep neural network parameterized by $\theta$, denoted as
$Q_\theta(\mathbf{s}_t, a_t)$, to help the agent maximize its cumulative reward from the environment. 
 To stabilize the training, a target network $Q_{\theta^\prime} (\mathbf{s}_t,a_t)$ is introduced, which is parameterized by 
$\theta^\prime$ and updated periodically by setting $\theta^\prime = \theta$. This delay in updating the target network enhances the stability of the original network $Q_\theta$.
With the experience replay buffer and the target network in place, the DQN algorithm minimizes the following loss function:
\begin{equation}
\label{eqn:q-learning}
\begin{split}
    L(\theta)  =  
    \EX_{(\mathbf{s}_t,a_t,\mathbf{s}_{t+1},r_t) \sim \mathcal{V}(B)} 
    [(r_{t} + \\
    \gamma * \max_{a_t} Q_{\theta^\prime}
    (\mathbf{s}_{t+1},a_t) - Q_{\theta}(\mathbf{s}_t,a_t))^2],
    % \EX_{(\mathbf{s}_t,a_t,\mathbf{s}_{t+1},r_t) \sim \mathcal{V}(B)} \[{(r_{t} + \\
    % \gamma * \max_{a_t} Q_{\theta^\prime}(\mathbf{s}_{t+1},a_t) - Q_{\theta}(\mathbf{s}_t,a_t))}^2\].
 \end{split}
\end{equation}
% Essentially, Q-learning aims to update a state-action value function $Q(\mathbf{s},a)$, that accepts a state \textbf{s} and an action $a$ and returns the value of taking that action given that state.
% To compute the Q-function, we let the agent explore the environment in many episodes to collect the rewards. 
% For each episode, at each time step $t$, we record the agent's current state $\mathbf{s_t}$, action $a_t$, the next state $\mathbf{s}_{t+1}$ and the reward $R_{t+1}$.
% Then we use classic Bellman equation [cite] to update the Q-function for each state-action pair:
% \begin{equation}
% \label{eqn:q-learning}
%     Q(\mathbf{s_t},a_t) = Q(\mathbf{s_t},a_t) + \alpha[R_{t+1} + \gamma * \max Q(\mathbf{s_{t+1}},a) - Q(\mathbf{s_t},a)]
% \end{equation}
% The inputs to the deep Q-network are the state $\mathbf{s_t}$ and action $a_t$, i.e., the Q-network represents a function $N(\mathbf{s_t},a_t)$.
% Its output is the value of that state-action pair, typically a real value.
% To train the deep Q-network, we minimize the loss function:
% \begin{equation}
%     L = (Q(\mathbf{s_t},a_t) - N(\mathbf{s_t},a_t))^2
% \end{equation}
where  $\lambda \in$ [0, 1) is the discount factor.
Mini batches of tuples $(\mathbf{s}_t,a_t,\mathbf{s}_{t+1},r_t)$ are sampled from the replay buffer $B$ with sample strategy $\mathcal{V}$, such as uniform distribution. 
Note that other extensions from DQN such as prioritized replay buffer \cite{schaul2015prioritized} or dueling networks \cite{wang2016dueling} can also be applied. %; but those are not used in our paper.

% In practice, the deep Q-network is usually designed so that, as proposed by DeepMind[cite], its input is the state and its output is the vector representing a distribution of the action values.

\subsection{Problem Formulation}
Given an agent $A$ characterized by its policy function $\mathcal{P}_A(\mathbf{s},a)$ and an original path $\tau_0 = <\mathbf{s}^{\tau_0}_1,\mathbf{s}^{\tau_0}_2,...,\mathbf{s}^{\tau_0}_{T_0}>$  made by $A$, our goal is to find a different path 
$\tau_r = <\mathbf{s}^{\tau_r}_1,\mathbf{s}^{\tau_r}_2,...,\mathbf{s}^{\tau_r}_{T_r}>
$ of sequence $\tau_0$ that achieves certain goal $\mathcal{G}$ while being {personalized} to agent $A$, at both path level and policy level.

\noindent\textbf{Goal} The goal $\mathcal{G}$ could represent different constraints or desired outcomes, depending on the task. For example, it could require agent $A$ to arrive at a particular destination state $\mathbf{s}_d$; or it could require the overall reward to be larger than some pre-defined threshold (e.g., a driver reaching the final destination faster); or in the context of sequence data in a supervised learning setting, it could require the sequence to be classified as a particular class. The goal can be formulated as a continuous outcome, the higher (lower) the better, or a binary outcome, whether certain criteria are met. % its counterfactual explanation has a certain desired label for sequence classification problems.

\noindent\textbf{Personalization at Path Level} Path-level personalization requires the $\tau_r$ to be as similar to the original path $\tau_0$  as possible.
For example, the new path $\tau_r$ recommended to the driver should have many overlapped states with the original path $\tau_0$ picked by the driver. 

Measuring the similarity (distance) between two sequences, $\tau_0$ and $\tau_r$, is a non-trivial task.
The problem becomes even more challenging if the sequences have different lengths.
For this reason, traditional metrics such as Euclidean or Manhattan distance metrics do not apply. 
% \textcolor{red}{use consistent notations, you use $\tau, \tau_0, \tau_r$ interchangeably. It's confusing}
In our work, we measure the distance between $\tau_0$ and any path $\tau_r$ by {Levenshtein distance} \cite{levenshtein1966binary}, i.e., the \textbf{edit distance}, denoted as $d(\tau_0,\tau_r)$, which counts the minimum number of operations (insertion, deletion, and substitution ) to convert $\tau_0$ to $\tau_r$.
For instance, the edit distance between the sequences "LRLRL" and "LRLRR" is 1 (only the last character differs), while the distance between "LRLRL" and "LRRLLR" is 3 (three characters need to be changed to make them identical).

% The Levenshtein distance, i.e., edit distance $d(\tau_0,\tau_r)$ between two sequences 
% $\tau_0 = <\mathbf{s}^{\tau_0}_1,\mathbf{s}^{\tau_0}_2,...,\mathbf{s}^{\tau_0}_{i}>$ and
% $\tau_r = <\mathbf{s}^{\tau_r}_1,\mathbf{s}^{\tau_r}_2,...,\mathbf{s}^{\tau_r}_{j}>$ with length $i$ and $j$ respectively is defined as:
% \[ 
% d_{ij}(\tau_0,\tau_r)  = \left\{
% \begin{array}{ll}
%       d_{i-1,j-1} & \mathrm{if} \:\: {\mathbf{s}_i^{\tau_0}}
      
%       = \mathbf{s}_j^{\tau_r} \\
%        \mathrm{min} (
%       d_{i-1,j} + w_{del}(\mathbf{s}_i^{\tau_0}), \\
%        d_{i,j-1} + w_{ins}(\mathbf{s}^{\tau_r}_j), \\
%       d_{i-1,j-1} + w_{sub}(\mathbf{s}_i^{\tau_0},\mathbf{s}^{\tau_r}_j) ) & \mathrm{if} \:\: \mathbf{s}_i^{\tau_0} \neq \mathbf{s}_j^{\tau_r}
% \end{array} 
% \right. 
% \]
% Here $w_{del}(\cdot),w_{ins}(\cdot)$ and $w_{sub}(\cdot, \cdot)$ are the functions returning the weighting scores of deletion, insertion, and substitution operation, respectively.
% In this paper, we set all those values to 1.

% \textcolor{red}{Can we use an example?}
%  \textcolor{red}{what is the definition of $S_\text{path}$ used in the experiments?}
% \textcolor{blue}{I provided the examples. $s_\text{path}$ is the path reward $R_\text{path}$ defined in section 4.1}
 
\noindent\textbf{ Personalization at Policy Level} This property requires the new path $\tau_r$ to align closely with the behavior of agent $A$. 
For example, if $\tau_r$ is a new path recommended to a driver aiming to reach a destination quickly, we might suggest a highway for an experienced driver and a local path for a novice driver. Essentially, the recommended path $\tau_r$ should respect agent $A$’s inherent behavior patterns.

In practice, the agent’s behavior can be provided by an external system or learned from a dedicated data stream, allowing us to either train a new policy using state-of-the-art techniques or directly use an existing one. In our method, we represent agent $A$’s behavior using the policy function $\mathcal{P}_A(.)$ \footnote{$\mathcal{P}_A$  can be given or estimated from data using many off the shelf methods such as imitation learning if given the agent's past demonstrations. There has been abundant work on policy estimation from data. To avoid distraction, we will not discuss how $\mathcal{P}_A$ can be built and assume it is given. }.

% \textcolor{blue}{Dat: Instead of differentiating personalization and similarity, we mention there are 2 types/levels of personalization: path-level and policy-level. \\
% Also, we shouldn't say we need the availability of $\mathcal{P}_A(.)$. 
% We should say that these days, there are a lot of pre-trained models and policies. This raises the question of how to incorporate them into training another policy that respects the behavior of those pre-trained models.
% }

% which is naturally reflected in its policy function $\mathcal{P}_A(\mathbf{s},a)$.

% \textcolor{black}{Note that personalization is different from similarity. 
% Personalization encourages the agent to follow a policy $\mathcal{P}_A$, which represents the expected behavior of the agent, i.e., what action/choice the agent will take given a state/situation. Therefore, personalization evaluates how much the generated path respects the distribution of behavior of the agent.
% %This knowledge is learned from the user by accumulating the data of user choices after a period of time. 
% On the other hand, the similarity is evaluated at the path level, which is specific to a given original path $\tau_0$. For example, an agent may have equal probabilities of taking the right path and left path from a crossroad. Suppose the original path $\tau_0$ goes right. Then the left path has a low similarity to $\tau_0$ but a high personalization score because the agent is equally likely to go left.}

\section{Method}
Our goal is to train a personalized recourse  policy
$\mathcal{\pi}_A^r$ for agent $A$ that can generate  recourse path $\tau_{r}$ that satisfies the three properties mentioned previously. To do that, we first design a reward function that encourages the policy $\mathcal{\pi}_A^r$ to incorporate the three properties. 
% We achieve this by designing a reward function that encourages the agent to 1) produce new path $\tau^\prime$ that is as similar as possible to $\tau$ 2) follow its behavior policy $\mathcal{P}_A$ and 3) achieve the counterfactual condition.

% Note that from $\mathcal{P}_A^r$, we can sample multiple counterfactual $\tau_{r}$ of $\tau$ to ensure $\textit{diversity}$ property.

\subsection{Reward Shaping}

\paragraph{Goal} First, a recourse path must satisfy  goals in the goal set $\mathcal{G}$.
It is important to note that the definition of this reward varies depending on the specific application. 
For instance, in some RL environments, the goal reward may refer to the total reward the agent can obtain from the environment, while in a supervised learning setting, the goal reward may correspond to the probability that the sequence is classified as positive by the model. 
Therefore, one should be able to define a goal reward $R_\text{goal}(\cdot)$ at the sequence level to encourage the agent to achieve the specified conditions within $\mathcal{G}$. 
Examples of how the goal reward is defined in various applications will be presented in the experiment section.

\paragraph{Path-level personalization} 
Given the edit distance $d(\tau_0,\tau_r)$ between two paths $\tau_0$ and $\tau_r$, as discussed in Section 3.2,  the similarity between $\tau_0$ and $\tau_r$ is computed by 
\begin{equation}
f_\text{sim}(\tau_0,\tau_r) = \frac{1}{d(\tau_0,\tau_r) + 1}.
\label{eq:f_sim}
\end{equation}
$f_\text{sim}(\cdot)$ value is between 0 and 1, where  $f_\text{sim}(\tau_0,\tau_r) = 1$ if $\tau_0$ and $\tau_r$ are identical, and close to 0 if they are very different.
$f_\text{sim}(\cdot)$ can be used as the path similarity reward $R_\text{path}$, which is defined at the sequence level.

\paragraph{Policy-level Personalization}
\label{section:personalization}
We aim to personalize the path according to the behavior policy \( \mathcal{P}_A \). This can be achieved by incorporating a personalization reward, \( R_{\text{policy}} \), into the reward function. This reward function is derived from agent \( A \)'s policy \( \mathcal{P}_A \), which is a probability distribution over state-action pairs. Specifically, for a given state-action pair \( (\mathbf{s}, a) \), \( \mathcal{P}_A \) returns a probability representing the likelihood that agent \( A \) will choose action \( a \) in state \( \mathbf{s} \). We utilize this probability to assess whether a recommended action \textit{aligns with} agent \( A \)’s behavior, as represented by this distribution.

Directly using the probability as the reward value (i.e., $h(p) = p$) could be misleading because the magnitude of the probabilities is a result of the cardinality of the action space. 
For example, if $|\mathcal{A}|=2$, a probability of 0.3 is considered low, and the personalization reward should also be low.
However, when $|\mathcal{A}|$ is a very large value, 0.3 should be considered large, and the corresponding personalization reward should also be large. Thus, a linear relationship between the reward and probability is insufficient.
Therefore, we design a link function 
to represent the reward
\begin{equation}
R_\text{policy}(\mathbf{s},a) = h(\mathcal{P}_A(\mathbf{s},a)).
\end{equation}

$h(\cdot)$  needs to satisfy the conditions as follows: 
\begin{enumerate}
\item  $h(p)$ increases monotonically with probability $p$.
\item $h(p)>0$ when $p > \frac{1}{|\mathcal{A}|}$, $h(p) = 0 $ when $p = \frac{1}{|\mathcal{A}|}$, and $h(p)<0$ when $p < \frac{1}{|\mathcal{A}|}$.
\item  When $p$ is close to 1, $h(p)$ is a large positive number and when $p$ is close to 0, $h(p)$ is a large negative number.
\end{enumerate}
In this paper, we design $h(\cdot)$ \footnote{See a visualization of $h(\cdot)$ in the supplementary material and justification for the design.
% \textcolor{red}{Dat you need to write a paragraph to show more justification of h function. Otherwise it is a very ad hoc function which will raise many questions from reviewers}
} as follows:
% \begin{equation}
%     h(p) = log(\frac{|\mathcal{A}|\cdot p}
%     {1-p}).
% \end{equation}
    \begin{equation}
            h(p)=
\begin{cases}
\log(\frac{p - \frac{2}{|\mathcal{A}|} + 1}
    {1-p}) 
 & \text{if $p >= \frac{1}{\mathcal{|A|}}$,} \\
\log(|\mathcal{A}| \cdot p) & 
\text{otherwise}.
\end{cases}
    \end{equation}

 % The $log()$ factors is to penalize a very large or very small value.
 % This encourages the agent to explore more.

Then, the total policy-level personalization reward defined at the sequence level is the sum of  rewards for each state-action pair along the path:
\begin{equation}
R_\text{policy} (\tau_r) = \sum_{\mathbf{s}_t \in \tau_r} R_\text{policy}(\mathbf{s}_t,a_t).
\end{equation}

Thus, the final reward for a path $\tau_r$ recoursed from $\tau_0$ has the form:
    \begin{equation}
    \label{eqn:total_reward}
    R(\tau_r) =  R_\text{goal}(\tau_r)+
    \lambda_\text{path} R_\text{path}(\tau_r,\tau_0) + 
    \lambda_\text{policy}  R_\text{policy}(\tau_r).
    \end{equation}
$\lambda_\text{path}$ and $\lambda_\text{policy}$ are parameters that determine the weighting of each reward component.
% They can be set to 0 if one
% It is important to note that all three components are not always necessary; the inclusion of each depends on the specific application and user preferences.

\subsection{Training}
We use DQN to train a recourse agent $\mathcal{\pi}_A^r$ that can generate a recourse path $\tau_r$ for $\tau_0$ executed by agent $\mathcal{P}_A$, such that $\tau_r$ maximizes the reward $R(\tau_r)$.
 For this paper, we employ the Upper Confidence Bound (UCB) bandit algorithm as our exploration strategy \cite{auer2002finite} but PPR is adaptable to different exploration strategies.

% We design an exploration strategy to help the training process converge faster. 

% \textcolor{red}{put t back in the formula}
% 
\paragraph{Exploration function}
We want to encourage the agent to explore states and actions with high state-action values, avoid repetition of the same states/actions, and pay higher priority to the new states/actions.
% We achieve those goals by using an idea that is similar to Upper Confidence Bound (UCB) bandit algorithm.
We define an exploration score for taking action $a_t$ from a state $\mathbf{s}_t$ at time $t$ as follows:

\begin{equation}
\label{eqn:ucb-exploration}
    E(\mathbf{s}_t,a_t) = Q_\theta(\mathbf{s}_t,a_t) + c_{e} \sqrt{\frac{\ln \:t}{N_t(\mathbf{s}_t,a_t)}}.
\end{equation}
Here, $Q_\theta(\mathbf{s}_t, a_t)$ is the state-action value. $N_t(\mathbf{s}_t,a_t)$ denotes the number of times action $a_t$ has been selected prior to time $t$ for state $\mathbf{s}_t$, and the number $c_{e} > 0$ controls the degree of exploration.
Intuitively, any state-action pair that has not been visited much before time $t$ will have a high $E(\mathbf{s}_t,a_t)$ score. 
During the exploration, at any state $\mathbf{s}_t$, the agent will give higher priority to the action that has a higher $E(\mathbf{s}_t,a_t)$ value.
In our experiments, we combine UCB and random exploration (or sample from personalized policy $\mathcal{P}_A$ for larger environments, e.g., text data) using $\epsilon$-greedy strategy with a decaying $\epsilon$. 

% \paragraph{Prioritized Experience Replay}
% \textcolor{black}{will be added with deep Q-network}

% Previous research [cite] shows that the correlation between all state-action pairs during the exploration steps can degrade the learning process.
% One proposed solution [cite] is first, we store all the experiences in a buffer called \textit{experience replay} during the exploration step.
% During the training process, we only take $k$ top experience from the buffer and shuffle them before training.
% This will make the training process become more effective.
% \subsection{The final algorithm}

\paragraph{Training algorithm}
The training algorithm is presented in Algorithm \ref{alg:alg1} to train $\pi_{A}^r$.
First, we explore the environment using the strategy in Equation (\ref{eqn:ucb-exploration}) and record all the states, actions as well as corresponding rewards (lines 1-9).
Next, we add the $k$ best records to buffer replay $B$ to train network $Q_\theta$ and update network $Q_\theta^\prime$ every $C$ step in the while loop (lines 10-16).
\textcolor{black}{Lines 12 converts the reward from trajectory-level to state-action level discounted return to update the network using Equation (\ref{eqn:q-learning})}.
This process is repeated until it converges, i.e., the agent achieve the optimal reward.

\RestyleAlgo{ruled}
%% This is needed if you want to add comments in
%% your algorithm with \Comment
\SetKwComment{Comment}{/* }{ */}

\LinesNumbered
\begin{algorithm}[hbt!]

\DontPrintSemicolon
\everypar={\nl}
% \begin{algorithmic}[1]
\KwIn{Original path $\tau_0$, personalized policy $\mathcal{P}_A$, an environment parameterized by state set $\mathcal{S}$, action set $\mathcal{A}$, and reward function $R(.)$}
\KwOut{A trained network $Q_\theta$ to sample $\tau_r$}

Initialize a replay buffer $B$ size $b$

Initialize two networks $Q_\theta, Q_{\theta^\prime}$ with random parameters $\theta$ and $\theta^\prime$

\While{not converge}
{
  $B = \emptyset$ 
  
  \For{i=1 \emph{\KwTo} $b$}{
  
    Sample a path $\tau = <\mathbf{s}_1,\mathbf{s}_2,...,\mathbf{s}_T>$ using Equation (\ref{eqn:ucb-exploration}) with decay $\epsilon$-greedy strategy

    Record all the action $a_t$ in the path

    Compute the reward $R(\tau)$ with Equation (\ref{eqn:total_reward})

    % Add ($\tau,R(\tau)$) to $B$
  }

  Collect $k$ best paths with the highest $R(\tau)$.

  \ForEach{$k$ record}{
    Compute the discounted return at every step $t$:
    \begin{equation*}
    r_t := R_{t} + \gamma R_{t+1} + \gamma^2 R_{t+2} + ... = R_{t} + \gamma * r_{t+1}
    \end{equation*}
    Add to $B$ all tuples $(\textbf{s}_t,a_t,\textbf{s}_{t+1},r_t)$
  }
    Samples tuples from $B$ and update $\theta$ for $Q_\theta$: 
    \begin{equation}
        \theta := \theta - \alpha * \Delta_{\theta} L(\theta)
    \end{equation}
    ($\alpha$ is the learning rate and $L$ is the loss function \ref{eqn:q-learning})
   
   Set $\theta^\prime = \theta$ every $C$ steps in the while loop
}

\caption{Train a personalized path recourse (PPR) agent}

\label{alg:alg1}
% \end{algorithmic}
\end{algorithm}

Note that PPR is compatible with various other training algorithms, including DQN, A3C, and PPO. In this paper, we use  DQN because it performs particularly well in environments with discrete action spaces and efficiently handles situations where interactions with the environment are limited \cite{de2024comparative,henderson2018deep}.
\subsection{PPR for Supervised Learning}
While PPR is motivated by and formulated in reinforcement learning, it
 can be applied to sequence data in supervised learning by aligning the properties and objectives with supervised learning. 
 
 In supervised learning, $\tau$ is not a sequence of states but a sequence data $\mathbf{x}$ with a label $y = Y_0$. A personalized recourse means to generate a new sequence $\mathbf{x}^\prime$ with a different label $Y_1$ (goal), while being similar to $\mathbf{x}$ (path-level) and being in-distribution (policy-level). 
 The goal reward can be defined as the probability that $\mathbf{x}^\prime$ is classified as label $Y_1$ by a pre-trained classifier. 
 The path-level personalization reward is defined the same way as in Section \ref{section:personalization}, which is the Levenshtein distance. However, other distance metrics such as Euclidean distance and Manhattan distance can also be used if the sequences have fixed lengths. 
 To incorporate path-level personalization reward, we train a generator model from the training data to represent agent $A$. For example, if it is text data, $\mathcal{P}_A$ is a language model; if it is a sequence of vectors, $\mathcal{P}_A$ can be a simple RNN model that is either pre-trained or provided. 
Then the personalization policy is defined as the sampling probability of the next symbol, given the current prefix sequence:
$   \mathcal{P}_A (\mathbf{s_t},a_t) = p(\mathbf{u_t}|\mathbf{u_1},...,\mathbf{u}_{t-1})$, where $\mathbf{u}_i$ represents each symbol in sequence data.

\section{Experiments}
In this section, we conduct experimental evaluations of PPR in several settings. In each setting, we generate \textbf{agents with different behavior} and then evaluate how PPR generates \emph{personalized}  recourse paths to adapt to those agents. The first two settings are reinforcement learning environments, the third one is text generation, and the last one is supervised learning. Due to the space limit, we included additional experiments and performance evaluations in the supplementary material. % The first is a reinforcement learning setting where we simulate two drivers with different driving habits and the goal is to get a higher total reward. The other two settings are for sequence data, where we work with time series and text data, respectively. The agents for these sequence data are generators trained from data.
%See the supplementary material for detailed information of each dataset.

\paragraph{Baselines} 
While we do not find any baselines that work for all settings or consider personalization, we find two relevant work that work for some settings but not all,  denoted as BL1 \cite{tsirtsis2021counterfactual} and BL2  \cite{delaney2021instance}, which work for some settings but not all.
BL1 focuses on generating an alternative sequence of actions differing in
at most $k$ actions from the observed sequence that could
have potentially a better outcome.
It
% % \textcolor{red}{add some explanation. k actions?}
% finds an alternative sequence of actions differing
% in at most k actions from the observed sequence, but it 
uses dynamic programming to exhaustively search all states and actions. 
Thus, it is not suitable for applications with very large state and action spaces, such as texts and atari games. BL2, on the other hand, is specifically designed to find counterfactual explanations for labeled sequences with the same lengths. 
It is used only for classification tasks in supervised learning settings and requires access to the training data and the classifier. Therefore, it is not suitable for reinforcement learning applications or text generation. Both implementation procedures can be found in the supplementary material.

\paragraph{Evaluation metrics} 
We evaluated the generated paths based on three key aspects: path-level personalization, policy-level personalization, and goal satisfaction, using corresponding scores $s_\text{policy}, s_\text{path},$ and $s_\text{goal}$, respectively. 
The goal satisfaction score $s_\text{goal}$ and path score $s_\text{path}$ are the same as the goal reward and the path-level reward, i.e., $s_\text{goal} = R_\text{goal}$ and $s_\text{path} = f_\text{sim}$ from Equation (\ref{eq:f_sim}).
%Note that similarity $f_\text{path}$ between PPR $\tau_r$ and the original path $\tau_0$ is computed by $f_\text{path}(\tau_0,\tau_r) = 1/(d + 1)$ where $d$ is the Levenshtein distance between $\tau_r$ and $\tau_0$.
% The higher those values the better.
For policy-level personalization score $s_\text{policy}$, we report the average log of probabilities of the generated PPR if it is sampled from the personalized policy $\mathcal{P}_A$, which represents the (normalized) likelihood of the new path - the higher this value, the  more PPR aligns with the  policy, i.e., the more likely the agent will take this path. % agent's behavior.
Thus, for a path $\tau_{r} = <\mathbf{s}_1^r,...,\mathbf{s}_T^r>$, 
    $s_\text{policy} = \frac{1}{T}  \sum_{t=1}^{T} \log \:
    \mathcal{P}_A(\mathbf{s}^r_{t+1} | \mathbf{s}^r_{t},a^r_t)$.
% \textcolor{red}{what is the definition of $S_\text{path}$. Also, we need to change the notation from $S_\text{path}$ to $S_\text{path}$}

% \textcolor{blue}{$s_\text{path}$ is $f_\text{sim}$.}
% \begin{enumerate}
%     \item (avg) personalization: probability (likelihood) of a new path
%     \item (avg) goal  (goal reward)
%     \item (avg) similarity (edit distance)
% \end{enumerate}
% \emph{Baselines}

% Baseline 1: a naive baseline where you find a sequence with the smallest distance and a different label
% Baseline 2: find some baselines from other papers and send a list to me. Let's discuss this. 
% 1. Counterfactual explanations in sequential decision making under uncertainty
% 2. Instance-based counterfactual explanations for time
% series classification. 

% conferences: ICLR, NeurIPS, ICML, AAAI, KDD
% try the full conference names and also initials

\subsection{Grid-world environment}
\label{sec:gridworld}
\begin{figure}
    \centering
        \includegraphics[width = 0.38\textwidth, height = 3.8cm]{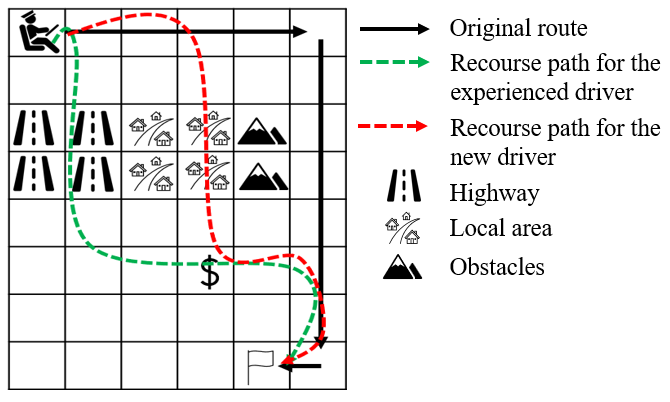}
    \caption{The original path 
    $\tau_0$ (the driver didn't collect reward), the recourse paths for the experienced driver $\tau_{\text{exp}}$  and the recourse path for the new driver $\tau_{\text{new}}$.}
    \label{fig:gridworld}
    \vspace{-5mm}
\end{figure}

% \textcolor{red}{generate 5 different original paths (plot all of them in the appendix), 2 agents, new and experienced, so in total generate 10 new paths. Results that we report. 1) plots of all new paths, 5 for each agent, 10 in total (appendix); 2) a table of 10 paths for new and experienced, report personalization, goal and similarity, together with a baseline. }
In this section, we apply PPR to a grid-world environment, shown in Figure \ref{fig:gridworld}.
In this environment, a taxi driver wants to go to the destination (the white flag) as soon as possible. If the driver reaches the destination, the reward is 80. 
An extra reward of 30  will be given if the driver can collect money at the dollar sign. The reward for any other step is -1.
Originally, the driver picked a bad path with a low reward, which either didn't collect the money or took a detour to reach the destination. We randomly generate 10 such bad paths and an example is shown in Figure \ref{fig:gridworld}. 
We want to generate a better path that can help the taxi driver get a better reward, which means she will reach the destination faster and collect the money. More importantly, we want to be able to provide \emph{personalized} recourse path that takes into consideration her driving habit. 
In addition, the new path should not be too different from the original one to satisfy the path level personalization. 
% \textcolor{red}{we need to mention we actually generated 20 (10?) different paths. So reviewers do not think Figure 1 is the only thing we did. Looks so simple and trivial}

% \vspace{-10mm}

We simulate two types of taxi drivers as agents. Driver \( A_\text{exp} \) represents an \textbf{experienced driver} who is comfortable driving on highways, while driver \( A_\text{new} \) is a \textbf{newbie driver} who prefers to avoid highways and take local roads instead. We model these two drivers by designing distinct reward functions. For agent \( A_\text{exp} \), we assign higher rewards for using highway routes and lower rewards for local roads, whereas for agent \( A_\text{new} \), the rewards are reversed\footnote{See the supplementary material for more detailed information.}. The policy functions derived from these agents are then provided as inputs to the PPR framework.

We then run Algorithm \ref{alg:alg1} to generate personalized improved paths for each agent. We set \( \lambda_\text{policy} = 0.1 \) and \( \lambda_\text{path} = 0.1 \) in Algorithm \ref{alg:alg1} (see the supplementary material for a sensitivity analysis of \( \lambda_\text{policy} \) and \( \lambda_\text{path} \)). An example is shown in Figure \ref{fig:gridworld}.

As depicted, the recommended path \( \tau_{\text{exp}} \), personalized for the experienced driver, passes through the highway area in accordance with the experienced driver’s policy. Similarly, the recommended path \( \tau_{\text{new}} \) for the newbie driver follows a local road, aligning with his preferences as reflected by the policy function. 

In this example, these new paths are the recourse paths to the original path, as they respect the driver's driving behavior, facilitate a better outcome for the taxi, and maintain similarity to the original path from the starting point to the destination. Please refer to the supplementary material for all 10 paths and their corresponding recourse paths.

\begin{figure}[h]
\centering
\begin{subfigure}{.45\textwidth}
  \centering  \includegraphics[width=0.95\linewidth,height=3cm]{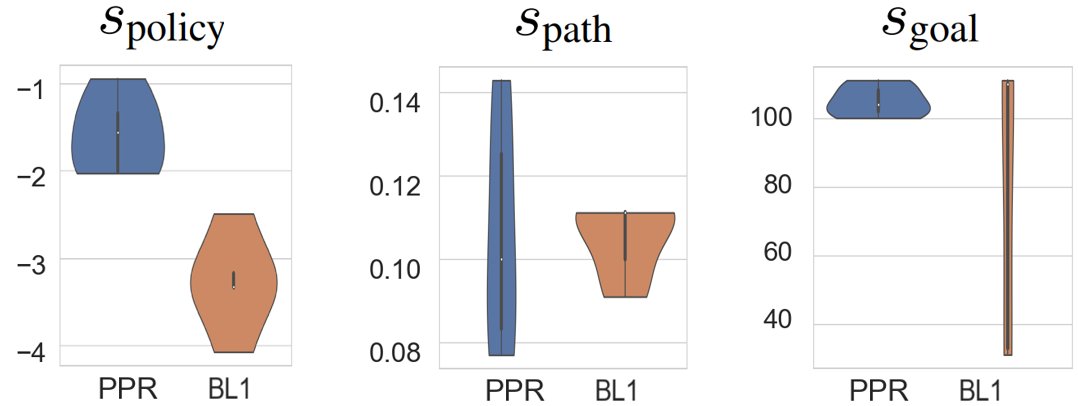}
  \caption{$s_\text{policy}$, $s_\text{path}$ ,$s_\text{goal}$ for PPR and BL1 trained on grid-world }
  % \label{fig:violin1}
\end{subfigure}%
\hfill
\begin{subfigure}{.45\textwidth}
  \centering
\vspace{1mm}
\includegraphics[width=1.01 \linewidth,height=3cm]{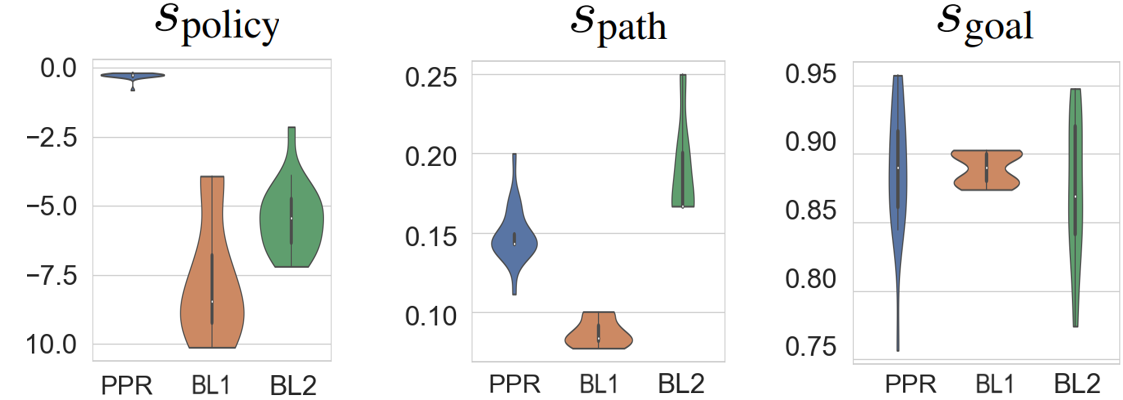}
  \caption{$s_\text{policy}$, $s_\text{path}$ ,$s_\text{goal}$ for PPR, BL1 and BL2 trained on temperature sequences (see the supplementary material for more details for this experiment).}
  % \label{fig:violin2}
\end{subfigure}
\caption{Personalization score $s_\text{policy}$, similarity score $s_\text{path}$, and goal score $s_\text{goal}$ of PPR and baselines BL1 and BL2. 
% \textcolor{red}{change y-axis range}
}
\label{fig:violin}
\end{figure}

We compare PPR with BL1 (BL2 is not applicable).
Figure \ref{fig:violin}(a) shows the violin plots of the three scores from PPR and BL1. Since BL1 works by changing $k$ actions at most, the similarity scores have a lower variance than PPR. However, both personalization and goal scores of PPR are significantly higher, especially the personalization score, which is not considered by BL1.
% \textcolor{red}{the subscripts in figure 2 should NOT be italic}

% % \usepackage{tabularray}
% \begin{table}
% \centering
% \caption{add some description here}
% \begin{tblr}{
%   cell{1}{1} = {c=2}{},
%   cell{2}{1} = {r=3}{},
%   cell{4}{3} = {c=10}{},
%   cell{5}{1} = {r=3}{},
%   cell{7}{3} = {c=10}{},
%   cell{8}{1} = {r=3}{},
%   cell{10}{3} = {c=10}{},
%   vlines,
%   hline{1-2,5,8,11} = {-}{},
%   hline{3-4,6-7,9-10} = {2-12}{},
% }
% Original path  &      & 1   & 2 & 3 & 4 & 5 & 6 & 7 & 8 & 9 & 10 \\
% Personalization & Ours &     &   &   &   &   &   &   &   &   &    \\
%                 & BL1  &     &   &   &   &   &   &   &   &   &    \\
%                 & BL2  & N/A &   &   &   &   &   &   &   &   &    \\
% Similarity      & Ours &     &   &   &   &   &   &   &   &   &    \\
%                 & BL1  &     &   &   &   &   &   &   &   &   &    \\
%                 & BL2  & N/A &   &   &   &   &   &   &   &   &    \\
% Goal reward     & Ours &     &   &   &   &   &   &   &   &   &    \\
%                 & BL1  &     &   &   &   &   &   &   &   &   &    \\
%                 & BL2  & N/A &   &   &   &   &   &   &   &   &    
% \end{tblr}
% \end{table}
% \vspace{-5mm}

\subsection{Mario Environment}
The second setting is the Mario environment provided by the OpenAI gym. This environment replicates the classic ``Super Mario Bros" game, where the player-controlled character, Mario, navigates through diverse levels filled with obstacles, enemies, and power-ups and the primary objective is to reach the end of each level. We work with level 1 and for simplicity, in our experiments, Mario can only take two actions for each state: move right or jump right. 
The game is ended by either winning by reaching the flag or losing by falling off a cliff or touching an enemy (e.g., a Goombas). 

In this environment, we simulate two types of players with different playing styles. The first player likes to collect coins when playing the game and we call him a \textbf{coin seeker}. To simulate a coin seeker, we provide an additional 100 whenever a coin is collected. The agent will then try to collect as many coins as possible by jumping at appropriate places. To contrast with the {coin seeker}, we simulate a different player who would avoid getting coins by giving a -100 reward whenever a coin is collected. We call the player \textbf{coin dodger}. The goal of simulating the two agents with the opposite behavior is to evaluate how that behavior is reflected in the recourse path. 

Then we construct an original path $\tau_0$ where Mario keeps moving right until getting killed by running into a goomba. The original path is generated by always choosing the action of ``go right", thus the path is featured by Mario walking in a straight line on the ground. 

\begin{figure}[h]
  \centering
  \includegraphics[width=0.5\textwidth]{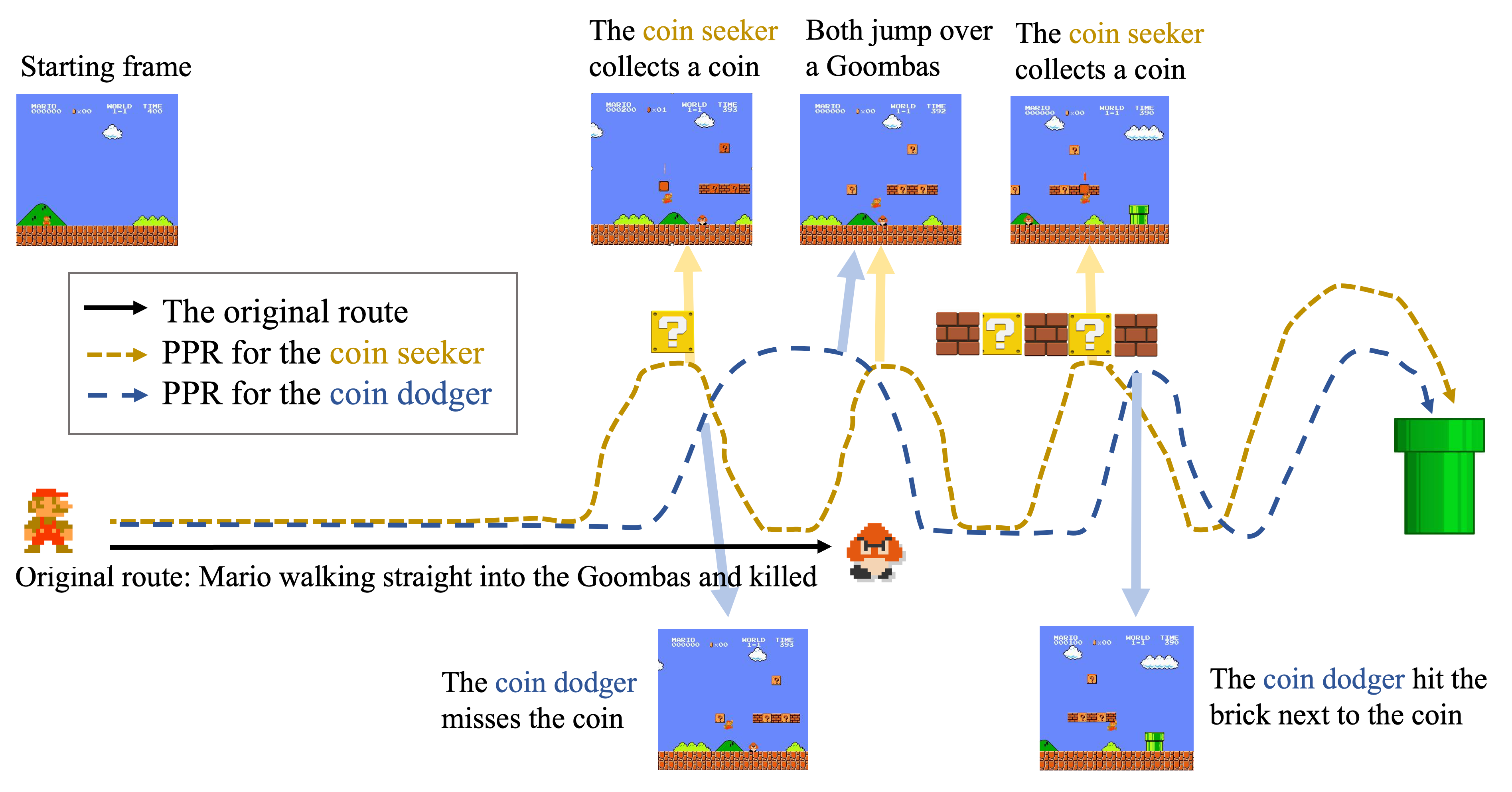}
  \caption{An illustration of one segment of level one of PPR generated for the coin seeker and coin dodger.}
  \label{fig:mario}
\end{figure}

Then we generate a recourse path for the coin seeker. We show a segment in Figure \ref{fig:mario} for demonstration. In the recourse path, the player successfully jumps over the Goombas to avoid dying and also collects two coins, one before Goombas and one after\footnote{In this segment there are a total of 3 coins that can be collected but the game is simplified to not allow going backward, then the player has to miss one coin if he wants to jump over a Goombas.}. 
We contrast this path with a recourse path generated for a coin dodger, also shown in Figure \ref{fig:mario}. As expected, the coin dodger did not hit any coins on the segment.

To quantitatively assess the extent to which the recourse path aligns with the preferences of the corresponding agent, we calculate the Kullback-Leibler (KL) divergence between the recoursed and original policy, shown in Table \ref{tab:KL}. To contrast with this ``personalized'' path, we also compare the policy of the PPR for the coin seeker with that of the coin dodger. In this case, while the generated path recourses the original path, it is not personalized to the coin dodger, thus significantly increasing the KL divergence from 0.342 to 1.049. We do the same for the coin seeker, comparing its policy function with the PPR generated for the coin dodger, and obtain a larger KL divergence of 0.554, up from 0.318 if it was personalized for him.
\begin{table}[h!]
\caption{The KL-divergence between agents' policy functions and the PPR}\label{tab:KL}
\footnotesize
\begin{tabular}{lll}
\toprule
            &   \textsc{Coin Seeker} &   \textsc{Coin Dodger} \\ \toprule
 \textsc{PPR for Coin Seeker }&    \textbf{0.318}                      &      1.049                  \\ \hline
 \textsc{PPR for Coin Dodger} &     0.554                        & \textbf{0.342}                 \\
\bottomrule
\end{tabular}
\end{table}

Note that baseline BL1 does not work for this experiment because it has to iterate through all the states of the environment, and the complexity of Mario environment is too high for BL1.
BL2 is not applicable for reinforcement learning.

\subsection{PPR for Text Data}
In this application, we apply the method to generate recourse sequences for text. The goal is to change the sentiment of an original text (goal) while tailoring it to a specific writing style that represents a specific agent (personalization at the policy level) and making as small changes as possible (personalization at the path level).

In this experiment, we train two language models as agents, an agent that represents the style of J.K. Rowling, trained from the 7  \textbf{Harry Potter} books, and another on the \textbf{Bible } corpus.
% More details of the description of the dataset and environment set-up can be found in the appendix. 
We employ the transformer to train our language models, which assigns probabilities for the likelihood of a given word or sequence of words following a given sequence of words. 
To evaluate the sentiment, we use a pre-trained sentiment analysis model, such as NLTK's sentiment library. 
This model returns a probability indicating the level of positive sentiment in the text. 
We use this sentiment score as the goal reward. 
It is worth noting that, PPR can directly work with an off-the-shelf language model and an evaluator (e.g., sentiment classifier), without accessing the training data, while many counterfactual explanation methods do need the training data to generate explanations. % except for the corpus to train the language model and the external pre-trained sentiment evaluation model, we do not need access to any text dataset. \textcolor{red}{let's discuss. why emphasizing this?}

To compare the recourse texts, we tune the hyperparameters by fixing $\lambda_\text{policy}$ to 1 and varying $\lambda_\text{path}$. 
Table 1 illustrates a few recourse texts generated from various original texts with different personalization policies and $\lambda_\text{path}$ values.
As shown in Table 1, the generated recourse sequences are personalized towards the training corpus and exhibit positive sentiment. 
The degree of similarity between the recourse text and the original text is controlled by the value of $\lambda_\text{path}$. 
Higher values of $\lambda_\text{path}$ result in recourse texts that are more similar to the original text, while lower values lead to more flexible recourse texts that still maintain a positive sentiment. See the supplementary material for more results. %It should be noted that the examples presented are merely a subset of the many paths that can be generated using PPR. 

Please note that neither BL1 nor BL2 can be applied to this application. BL1 faces practical challenges because of its high computational complexity of $\mathcal{O}(n^2 mTk)$ ($n,m, T,k$ are the number of states, the number of actions, sequence length and the number of actions changed, respectively). For text data, both states and action space are extremely large, making BL1 computationally infeasible and creating a timeout error. BL2 is also not applicable due to the unavailability of a labeled text dataset, while PPR can directly work with a trained sentiment classifier.

\begin{table}[h!]
  \centering
  \caption{Examples of text recourse with different writing styles and $\lambda_\text{path}$. }
  \label{tab:text-cf}
  \footnotesize
  \renewcommand{\arraystretch}{0.4}
  \begin{tabular}{p{8cm}}
    \toprule
    \textsc{Original text:} She was sad \\
    \midrule
    \textbf{$\mathcal{P}_A$ trained from Harry Potter corpus:}
    \begin{itemize}[noitemsep, topsep=0pt, parsep=0pt]
        \item $\lambda_\text{path} = 10$: She was very happy.
        \item $\lambda_\text{path} = 0.1$: She giggled maliciously.
    \end{itemize}
    \textbf{$\mathcal{P}_A$ trained from Bible corpus:}
    \begin{itemize}[noitemsep, topsep=0pt, parsep=0pt]
        \item $\lambda_\text{path} = 10$: She was exceedingly beautiful.
        \item $\lambda_\text{path} = 0.1$: She shall glorify God and honor and she shall be known.
    \end{itemize}\\
    \midrule
    \textsc{Original text:} The cup was empty. \\
    \midrule
    \textbf{$\mathcal{P}_A$ trained from Harry Potter corpus:}
    \begin{itemize}[noitemsep, topsep=0pt, parsep=0pt]
        \item $\lambda_\text{path} = 10$: The cup was very strong.
        \item $\lambda_\text{path} = 0.1$: The cup and a strong love potion again.
    \end{itemize}
    \textbf{$\mathcal{P}_A$ trained from Bible corpus:}
    \begin{itemize}[noitemsep, topsep=0pt, parsep=0pt]
        \item $\lambda_\text{path} = 10$: The cup was altogether lovely.
        \item $\lambda_\text{path} = 0.1$: The cup is of the Lord, thy God with a blessing.
    \end{itemize}\\
    \midrule
    \textsc{Original text:} The book on the table is boring. \\
    \midrule
    \textbf{$\mathcal{P}_A$ trained from Harry Potter corpus:}
    \begin{itemize}[noitemsep, topsep=0pt, parsep=0pt]
        \item $\lambda_\text{path} = 10$: The book on the wall and the other hand were completely invisible.
        \item $\lambda_\text{path} = 0.1$: The book of interesting facts and Harry had a very enjoyable morning.
    \end{itemize}
    \textbf{$\mathcal{P}_A$ trained from Bible corpus:}
    \begin{itemize}[noitemsep, topsep=0pt, parsep=0pt]
        \item $\lambda_\text{path} = 10$: The book of Jashar, thy glory, which is exalted at the right hand of God.
        \item $\lambda_\text{path} = 0.1$: The book of Moses blessed the Lord and the God of Israel.
    \end{itemize}\\
    \bottomrule
  \end{tabular}
\end{table}

\subsection{Other Experiments}
We include more experiments in the supplementary material.

\paragraph{Extension to recourse of supervised models} As discussed in Section 4.3, PPR can also be applied to supervised learning settings. We use PPR to generate recourse sequences for yearly temperature patterns from a dataset with 200 years of yearly temperatures for  Rome in Italy and Columbia in South Carolina, USA. In this experiment, we want to recourse the temperature sequences from one city to the other while ensuring the generated sequence is similar to the original temperate (path similarity) and still follows the pattern and style of the temperature patterns in the data (policy-level personalization). Results are shown in Figure \ref{fig:violin} (b).

\begin{figure}
\centering
% \vspace{-6mm}
\begin{minipage}{\linewidth}
\centerline{\includegraphics[width=5cm, height = 2.8cm]{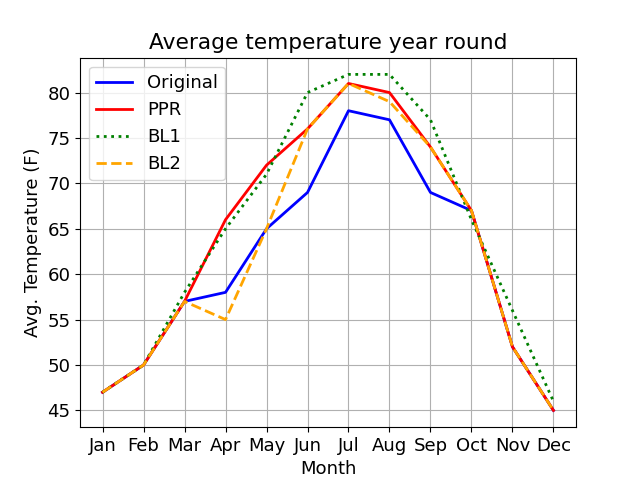}}
  \subcaption{recourse from Rome to Columbia}
  \label{fig:temp1}
\centerline{\includegraphics[width=5cm, height = 2.8cm]{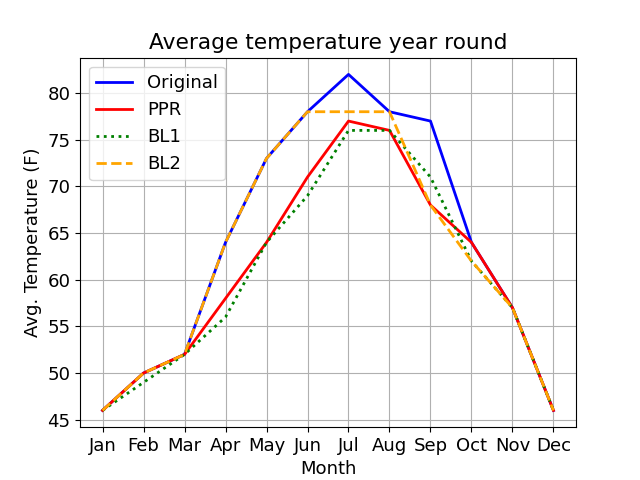}}
  \subcaption{  recourse from Columbia to Rome}
  \label{fig:temp2}
% \label{fig:violin}
% \vspace{-8mm}
\end{minipage}
\caption{Examples of recourse temperate generated by different methods}
\vspace{-5mm}
\label{fig:temperature}
\end{figure}

For example, to recourse a Rome temperature sequence to Columbia (goal), we begin by training a generator model on all temperature data, which acts as the personalized policy for our algorithm. 
% The generator model is trained with standard LSTM architecture.
Next, we train a classifier on our dataset, and the classification probability score serves as the goal reward, which has a higher score indicating a higher chance that the temperature sequence belongs to Columbia. 
We use standard LSTM architecture for both the generator model and the classifier.
Finally, we execute Algorithm \ref{alg:alg1} to generate  recourse sequences. % Columbia temperature sequence, which is similar to the input sequence from Rome. 
%Conversely, to generate a Rome sequence pattern that is similar to Columbia's, the process is repeated again but the dataset labels are flipped. 
Figure \ref{fig:temperature} presents the given original temperature sequences and the corresponding recourse sequences generated from PPR.
We also show the sequences generated by BL1 and BL2 for comparison.

Our recourse sequences show a high degree of similarity to the original temperature sequences. 
For example, in Figure \ref{fig:temperature}(a), the recourse sequences accurately reflect the temperature pattern in Columbia while closely resembling the temperature pattern of Rome during fall and winter months (from October to March). 
The recourse sequences reveal that the main differences between the two cities' temperature patterns lie in the summer months.

% \textcolor{red}{increase the fontsize of the labels and titles in the figure. They should have the same size as the text in the paper}
% \begin{figure}[h]
% \centering
% \includegraphics[scale=0.5]{imgs/temperature.png}
% \caption{Input sequence and the recourse sequence \textcolor{red}{4 figures, 2 for each class recourse to the other class, arrange them in 1 by 4}}
% \label{fig:temperature}
% \end{figure}

% \textcolor{red}{keep 2 subfigures only in FIgure 3}
% The PPR sequence generated for the input Rome's temperature sequence is depicted by the red line, and 

% \textit{Sensitivity Analysis:} Similar to the previous section, we also do sensitivity analysis by evaluating the effect of different hyper-parameter values from equation \ref{eqn:total_reward} on PPR quality.
% In this application, we randomly select 10 Columbia temperature sequences and 10 Rome temperature sequences.
% We then set different values $10e^{-1},10e^0,10e^1,10e^2$ for $\lambda_\text{policy},\lambda_\text{path}$ to generate the PPRs, and compute their $s_\text{policy}, s_\text{path}$ and $s_\text{goal}$. 
% We report the average scores for each combination of hyper-parameter values in Figure \ref{fig:app2_heatmap}.

Figure \ref{fig:violin}(b) compares $s_\text{policy},s_\text{path}$ and $s_\text{goal}$ of PPR with BL1 and BL2. 
Overall, PPR achieves significantly higher $s_\text{policy}$ scores  while maintaining  $s_\text{goal}$ scores. Having low $s_\text{policy}$ indicates that some of the sequences generated by BL1 and BL2 deviate from the data distribution. 
For instance, BL2 shows April is colder than March in Figure \ref{fig:temperature}(a), and  all three summer months have the same temperatures in Figure \ref{fig:temperature}(b) -- these are inconsistent with the patterns in training data, indicating the generated sequence is out of distribution.
BL1 has a lower similarity to the original temperature - the recourse sequence deviates from the original sequence more than necessary. As shown in Figure \ref{fig:temperature} (a) and (b), PPR only needs to change the summer temperatures, leaving other months the same, while BL1 changes the temperature of almost every month, though smaller changes in winter.
% For instance, as shown in Figure (b), the BL2 sequence shows
% April as being colder than March, which is inconsistent with the data.
In addition, BL1 has high variance in terms of the personalization score, as it exhaustively searches all possible states in the dataset.

\paragraph{Influence of $\lambda_\text{policy}$ and $\lambda_\text{path}$} We also investigate the influence of $\lambda_\text{policy}$ and $\lambda_\text{path}$ from equation (\ref{eqn:total_reward}) on recourse path quality.
For this type of experiment,  we set different values for $\lambda_\text{policy}$, $\lambda_\text{path}$, and report the average $s_\text{policy},s_\text{path}$ and $s_\text{goal}$ of the recourse paths in the generated recourse path set.
The details of the experiment setup and results can be found in the supplementary material.
In general, the higher values of $\lambda_\text{policy}$ and $\lambda_\text{path}$ are, the higher $s_\text{policy}$ and $s_\text{path}$ obtained.
When both of these values are low, the recourse paths tend to have higher goal reward $s_\text{goal}$.

\section{Conclusion}
We introduced Personalized Path Recourse (PPR), which generates personalized paths of actions that achieve a certain goal (e.g., a better outcome) for an agent. 
Our approach extends the existing literature on counterfactual explanations and generates entire personalized paths of actions, rather than just explaining why an action is chosen over others. Existing baselines either do not work for applications with large state and action space, or can only be applied to supervised learning. PPR, on the other hand, can be applied to a broader set of applications.
% PPR can potentially be applied to a variety of fields, including personalized recommendation systems, policy optimization in reinforcement learning, and sequence data modification in supervised learning. 

% \paragraph{Societal Impact}

\paragraph{Limitations} 
First, the current PPR approach is limited to discrete states or sequences. Future work could explore extending PPR to more complex decision-making scenarios that involve continuous states and diverse data modalities. Another limitation is the dependency on the availability of an agent's policy, which must be either provided or trained. Therefore, future research could investigate the application of PPR with limited training data. However, since there is extensive research on policy estimation from limited data, this aspect is beyond the scope of this paper.

\bibliography{aaai24}

\newpage
\appendix
\onecolumn
\pagenumbering{gobble}
\section{Experiment design} 

All the codes for the experiments are uploaded in the supplementary materials with the instruction .txt file.
They are implemented in Python 3.9 and Pytorch 1.13.
All the models are trained on GPU NVIDIA GeForce GTX 1660 Ti.
The training process uses AdamOptimizer \url{https://pytorch.org/docs/stable/generated/torch.optim.Adam.html}.
\subsection{Grid-world}
\paragraph{Experiment set-up} In this experiment, we create a 8x6 array and set $b=200,k=1,c_{e}=1,\gamma = 0.99,\lambda_\text{policy}=0.1,\lambda_\text{path}=0.1,\text{decay} \: \epsilon=0.001,C=1$ to run the Algorithm 1.
The learning rate was $1e-3$
% First, we choose one original route to generate the result in Figure 1 of the main paper.
We select 10 random original routes and generate 10 corresponding recourse paths.
We report one example in Figure \ref{fig:gridworld} in the main paper.

\begin{figure}[htbp]
\centering
\begin{subfigure}{.2\textwidth}
    \centering
    \includegraphics[width=.95\linewidth]{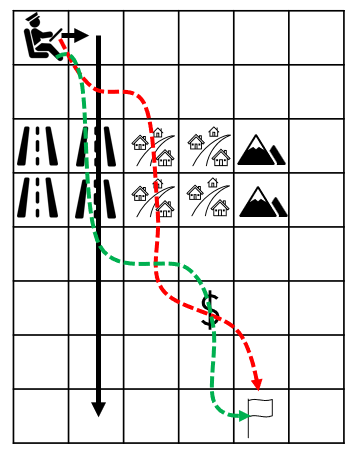}  
    \caption{}
    \label{fig:figure14_1}
\end{subfigure}
\begin{subfigure}{.2\textwidth}
    \centering
    \includegraphics[width=.95\linewidth]{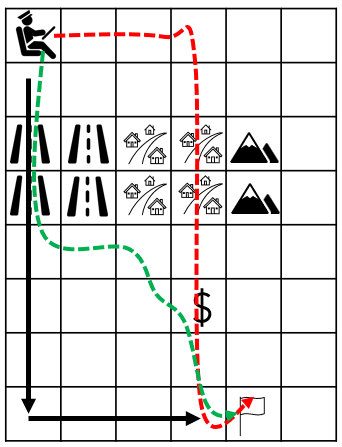}  
    \caption{}
    \label{fig:figure14_2}
\end{subfigure}
\begin{subfigure}{.2\textwidth}
    \centering
    \includegraphics[width=.95\linewidth]{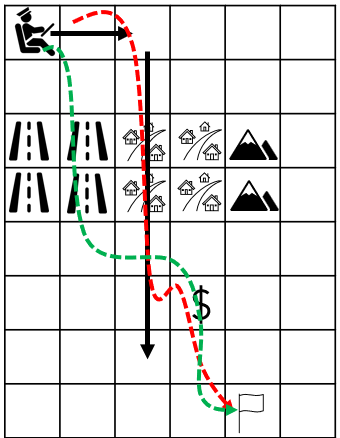}  
    \caption{}
    \label{fig:figure14_3}
\end{subfigure}
\begin{subfigure}{.2\textwidth}
    \centering
    \includegraphics[width=.95\linewidth]{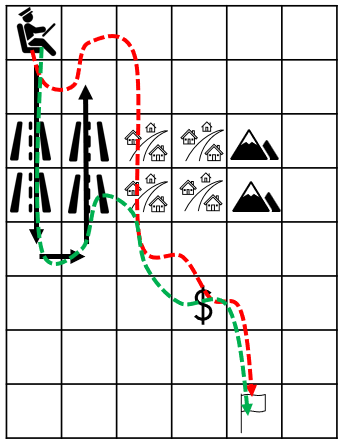}  
    \caption{}
    \label{fig:figure14_4}
\end{subfigure}

\begin{subfigure}{.2\textwidth}
    \centering
    \includegraphics[width=.95\linewidth]{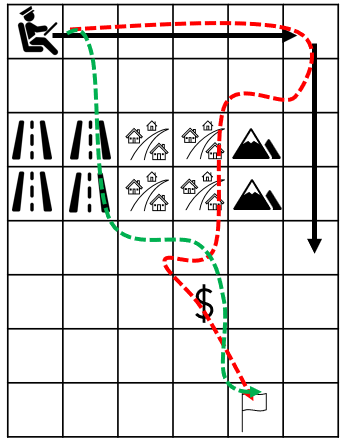}  
    \caption{}
    \label{fig:figure14_5}
\end{subfigure}
\begin{subfigure}{.2\textwidth}
    \centering
    \includegraphics[width=.95\linewidth]{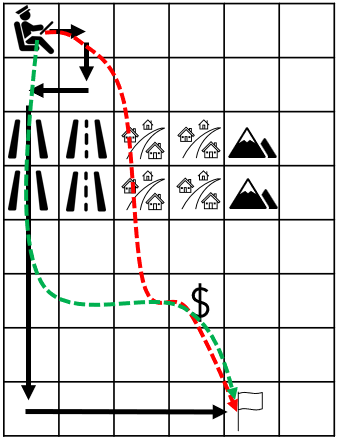}  
    \caption{}
    \label{fig:figure14_6}
\end{subfigure}
\begin{subfigure}{.2\textwidth}
    \centering
    \includegraphics[width=.95\linewidth]{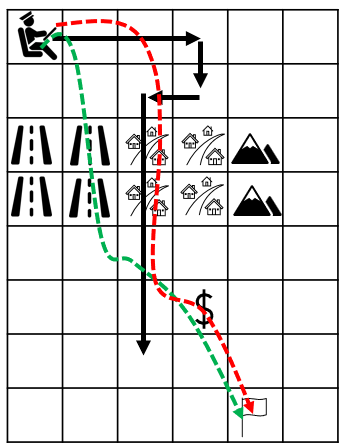}  
    \caption{}
    \label{fig:figure14_7}
\end{subfigure}
\begin{subfigure}{.2\textwidth}
    \centering
    \includegraphics[width=.95\linewidth]{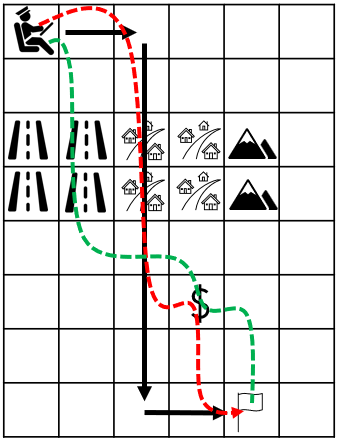}  
    \caption{}
    \label{fig:figure14_8}
\end{subfigure}

\caption{Other random original routes and the corresponding recourse paths, in addition to Figure \ref{fig:gridworld}.}
\label{fig:gridworld_allroutes}
\end{figure}

\paragraph{Original and recoursed paths}
In addition to the example in the main paper, the other 8 original routes and 
the corresponding recourse paths trained on experienced driver policy are shown in Figure \ref{fig:gridworld_allroutes}.
 We use those results to generate the plots in Figure \ref{fig:violin}(a) in the main paper.

\begin{figure}[h]
     \centering
     \begin{subfigure}[b]{0.3\textwidth}
         \centering
         \includegraphics[height = 3.5 cm, width=\textwidth]{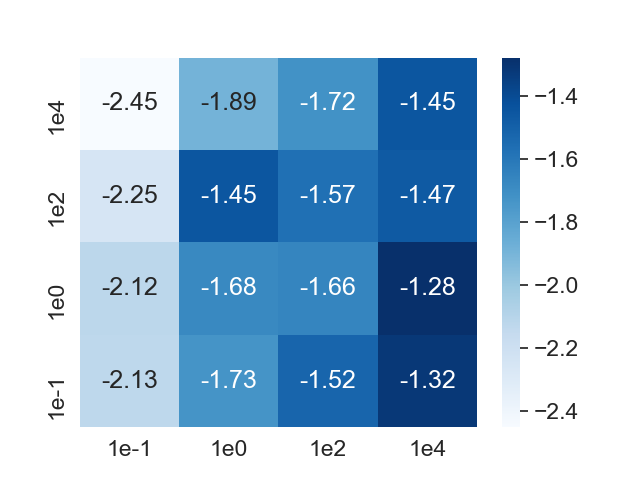}
         \caption{Personalization $s_\text{policy}$}
         \label{fig:gridworld_original}
     \end{subfigure}
     \hfill
     \begin{subfigure}[b]{0.3\textwidth}
         \centering
         \includegraphics[height = 3.5 cm, width=\textwidth]{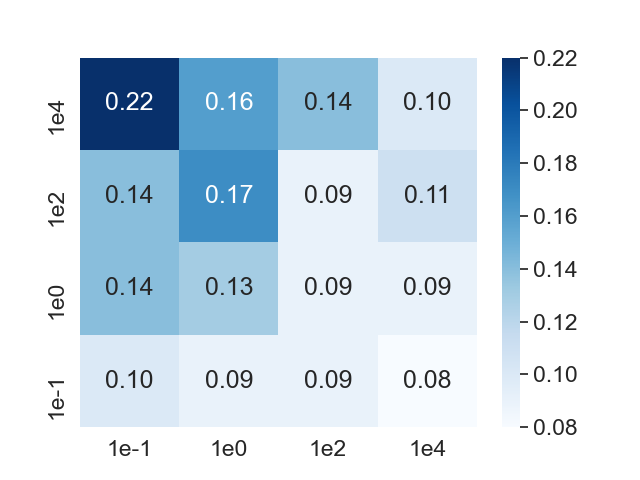}
         \caption{Similarity $s_\text{path}$}
         \label{fig:gridworld_experienced}
     \end{subfigure}
     \hfill
     \begin{subfigure}[b]{ 0.3\textwidth}
         \centering
         \includegraphics[height = 3.5 cm, width =\textwidth]{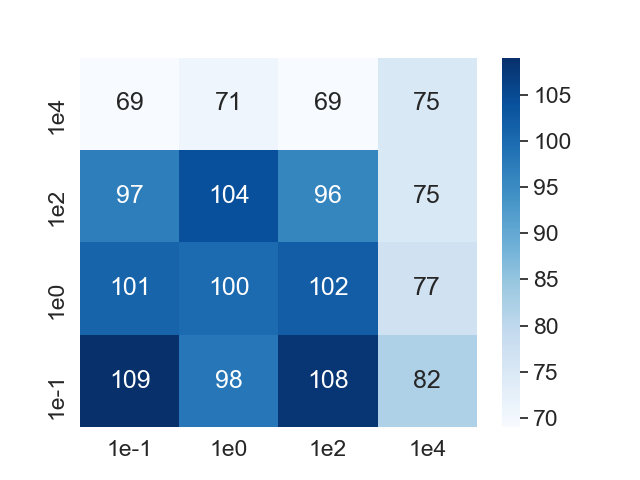}
         \caption{Goal score $s_\text{goal}$}
         \label{fig:five over x}
     \end{subfigure}
        \caption{Personalization, similarity and goal score $s_\text{policy},s_\text{path},s_\text{goal}$ for the grid-world experiment. The x-axis indicates different values of $\lambda_\text{path}$ and the y-axis shows different values of $\lambda_\text{policy}$.
        }
        \label{fig:heatmap_gridworld}
\end{figure}

\paragraph{Sensitivity Analysis} We conduct sensitivity analysis using the same 10 original routes but with different values of $\lambda_\text{policy}, \lambda_\text{path}$.% to different values such as $1e^-1,1e^0,1e^2,1e^4$.
We report the heatmaps of  the average scores of $s_\text{policy},s_\text{path},s_\text{goal}$ in Figure 
\ref{fig:heatmap_gridworld}.
In general, the higher values of $\lambda_\text{policy}$ and $\lambda_\text{path}$ are, the higher $s_\text{policy}$ and $s_\text{path}$ we can get.
When both of these values are low, such as 0.01, the PPRs tend to have higher goal reward $s_\text{goal}$.

% \paragraph{Baseline comparison:} For BL1, we set $k$ empirically and report the best value.
% Note that BL2 is not applicable in this 

\paragraph{Baselines}
As a reminder, BL2 is not applicable because it is designed only for classification tasks in supervised learning settings.
We are comparing our method PPR to BL1 in this application.
We implement BL1 by exhaustively searching all possible states and actions of the environment and recording all the valid paths (the ones that do not fall off the grid nor pass through the obstacles).
The path with the maximum goal reward is reported.
$k$ is set empirically for BL1.
Note that BL1 aims to find a recourse path that has $k$ different actions from the original path while maintaining the same length. 
Therefore, when the original path is too short, recourse paths generated from BL1 may not reach the final destination.
This is one of the reasons why its goal reward is low in some cases.

\subsection{Mario Game}

In this paper, we implement PPR based on the Mario Reinforcement Learning tutorial from 
\url{https://pytorch.org/tutorials/intermediate/mario\_rl\_tutorial.html} by Yuansong Feng, Suraj Subramanian, Howard Wang, and Steven Guo.
The tutorial gives instructions on how to train Mario agent to play itself by using Double Deep Q-Networks) using PyTorch.
In this tutorial, Mario's action space is limited to only two actions: Walk Right or Jump Right.
The real state is created by stacking up 3 consecutive game frames, which are also reduced to grayscale.
For all the experiments, we train Mario on the first stage/mission only.

The Mario environment, powered by Open AI gym \texttt{gym-super-mario-bros}, provided the \texttt{info} variable to help users collect the information from the environment such as how many coins Mario collected, how many scores Mario has had, the (x,y) coordinate of the agent's current position, how many lives remaining, etc.
By default, the environment reward considers three factors: the difference in agent x values between states, the difference in the game clock between frames, and a death penalty that penalizes the agent for dying in a state.
The environment reward is capped to a [-15,15] range.
In our experiments, this reward is used as the goal reward.
To train a policy to encourage the agent to collect coins, we alter the environment by giving an extra +100 reward whenever Mario collects a coin. 
Similarly, a reward of -100 will be taken from the agent to discourage it from collecting coins.
% We keep all the parameters default when training the personalization policies.

\begin{figure}[h]
     \centering
     \begin{subfigure}[b]{0.7 \textwidth}
         \centering
         \includegraphics[height = 1.5 cm, width=\textwidth]{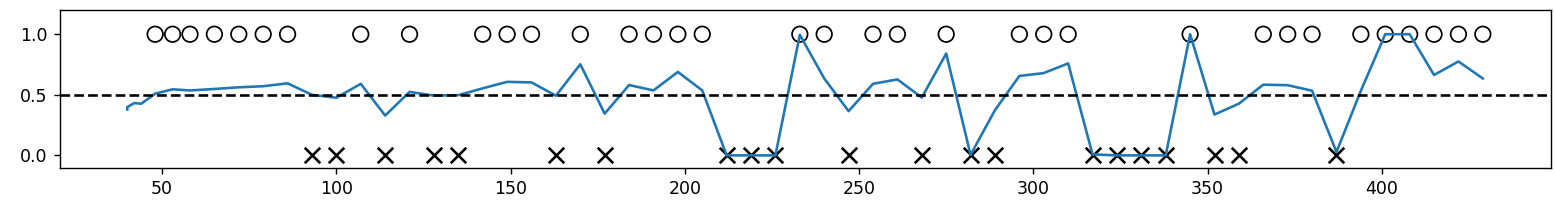}
         \caption{Coin seeker}
         % \label{fig:gridworld_original}
     \end{subfigure}
     \hfill
     \begin{subfigure}[b]{0.7\textwidth}
         \centering
         \includegraphics[height = 1.5 cm, width=\textwidth]{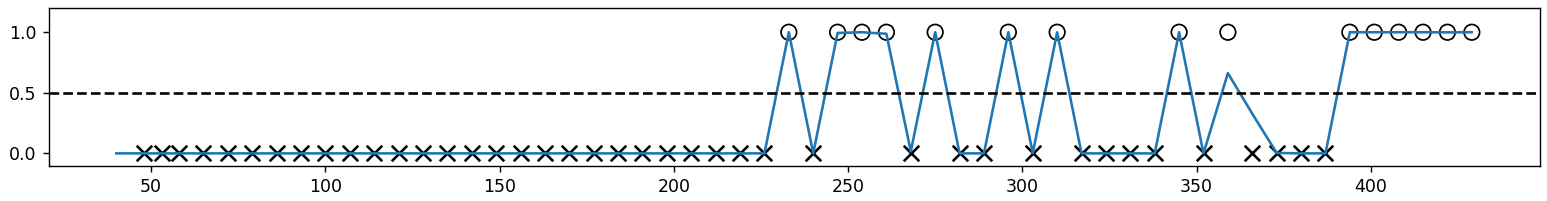}
         \caption{PPR for coin seeker}
         % \label{fig:gridworld_experienced}
     \end{subfigure}
     \hfill
     \begin{subfigure}[b]{ 0.7\textwidth}
         \centering
         \includegraphics[height = 1.5 cm, width =\textwidth]{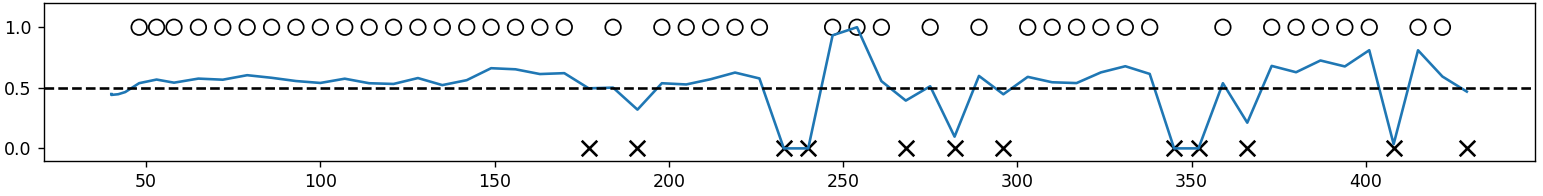}
         \caption{Coin dodger}
         % \label{fig:five over x}
     \end{subfigure}
     \hfill
     \begin{subfigure}[b]{ 0.7\textwidth}
         \centering
         \includegraphics[height = 1.5 cm, width =\textwidth]{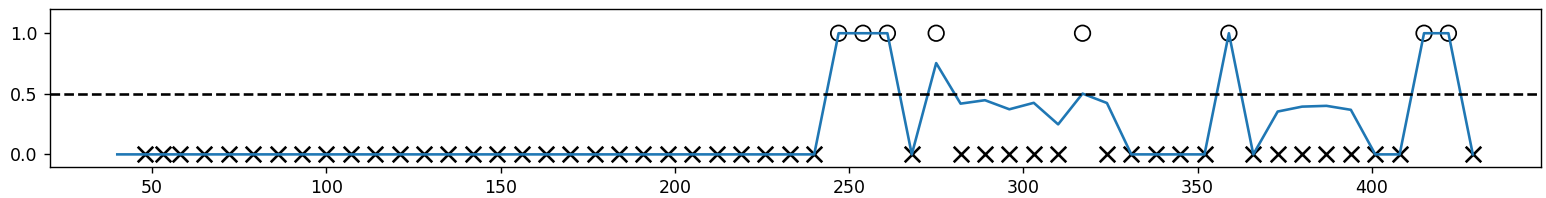}
         \caption{PPR for coin dodger}
         % \label{fig:five over x}
     \end{subfigure}
        \caption{Probability distributions of action "Jump Right" at each position. 'O' indicates where Mario takes ``Jump'' actions and 'X' indicates where Mario takes 'Walk' (``no jump'') actions.}
        \label{fig:mario_distribution}
\end{figure}

To implement PPR, we set $\lambda_\text{path} = 100$ and $\lambda_\text{route} = 10$.
% The original route is set as the straight line from Mario's starting position to right before the first brick which contains the coin (the brick with the question mark ``?'').
All the policies are trained with at least 50,000 episodes. 
Figure \ref{fig:mario_distribution} shows the probability distribution of taking action "Jump Right" for all policies: coin seeker, coin dodger, and their PPR policies for the first 460 steps (x-coordinate).

\subsection{Temperature}

\textbf{Dataset} In this experiment, we use the temperature dataset from Kaggle \url{https://www.kaggle.com/datasets/sudalairajkumar/daily-temperature-of-major-cities}, which contains daily average temperature values recorded in major cities of the world.
We choose 2 cities Rome from Italy and Columbia from North Carolina, USA.
For each city, we compute the average monthly temperature for each year from 1995 to 2020.
This results in 25 temperature sequences with lengths of 12 for each city.
We also apply data augmentation by randomly adding or subtracting randomly 1-2 degrees from each month, while maintaining the order pattern of each sequence. The label for each sequence is the city (either Rome or Columbia).
We choose 200 temperature sequences for the training dataset and 20 sequences for the test dataset (10 temperature sequences for Rome and 10 temperature sequences for Columbia). 
% The list of all those temperature sequences are the recourse temperature sequences generated by PPR and BL1, BL2 are shown in Figure \ref{fig:temperature2}.

\paragraph{Experiment set-up} To run  Algorithm 1 in the main paper, we set $b=500,k=1,\gamma = 0.99,c_{e}=1,\lambda_\text{policy}=0.1,\lambda_\text{path}=10,\text{decay} \: \epsilon=0.001,C=1$.
The learning rate value is $1e-3$.
We train the generator and the classifier using the standard LSTM architecture.

\paragraph{Original sequences and recoursed sequences}
Figure \ref{fig:temperature2} and \ref{fig:temperature3} show the 20 original temperature sequences from the test set and the corresponding recourse paths generated.

\begin{figure}[htbp]
\begin{subfigure}[t]{0.3\textwidth}
    \includegraphics[width=\linewidth]{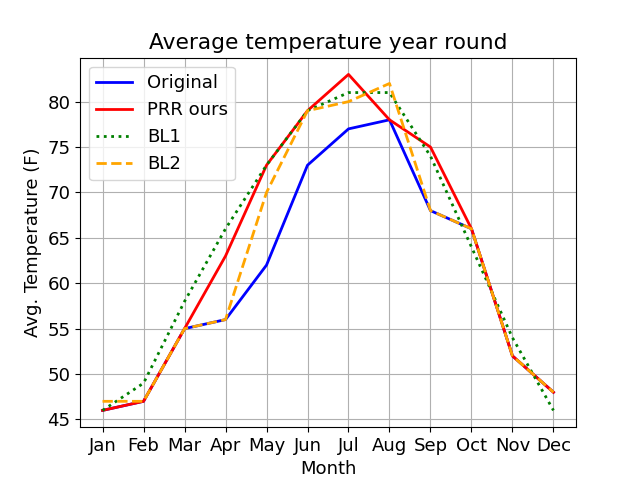}
\caption{}
\label{fig:figure14_1}
\end{subfigure}\hfill
\begin{subfigure}[t]{0.3\textwidth}
  \includegraphics[width=\linewidth]{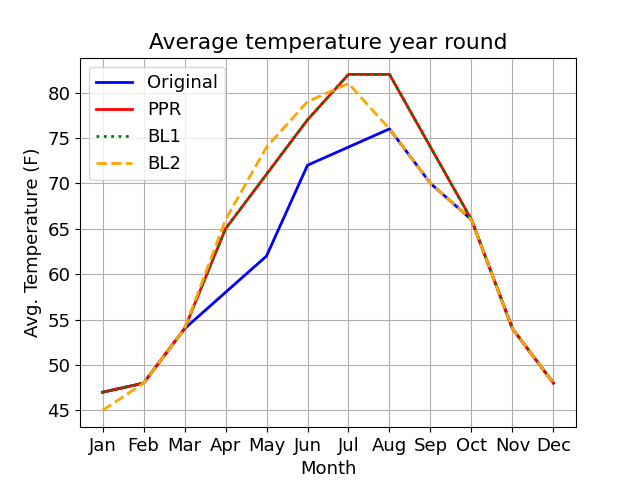}
\caption{}
\label{fig:figure14_2}
\end{subfigure}\hfill
\begin{subfigure}[t]{0.3\textwidth}
    \includegraphics[width=\linewidth]{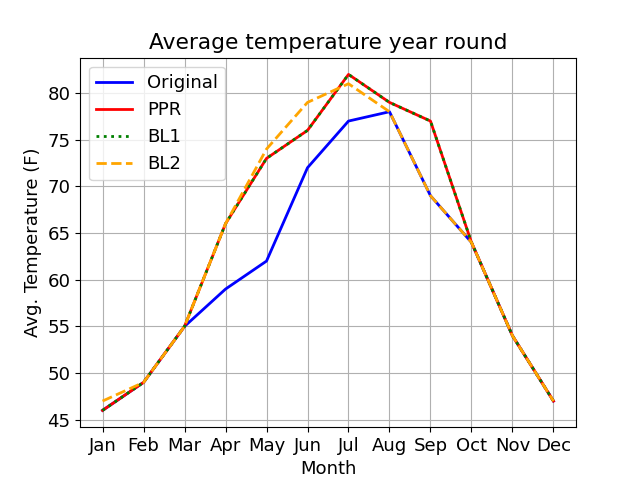}
\caption{}
\label{fig:figure14_3}
\end{subfigure}

\begin{subfigure}[t]{0.3\textwidth}
    \includegraphics[width=\linewidth]{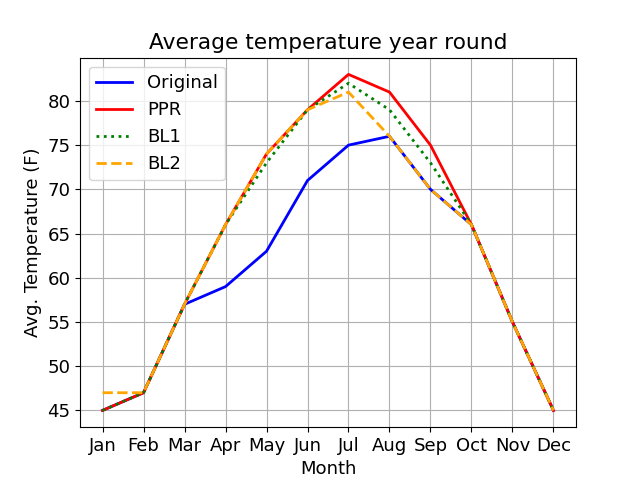}
\caption{}
\label{fig:figure14_4}
\end{subfigure}\hfill
\begin{subfigure}[t]{0.3\textwidth}
    \includegraphics[width=\linewidth]{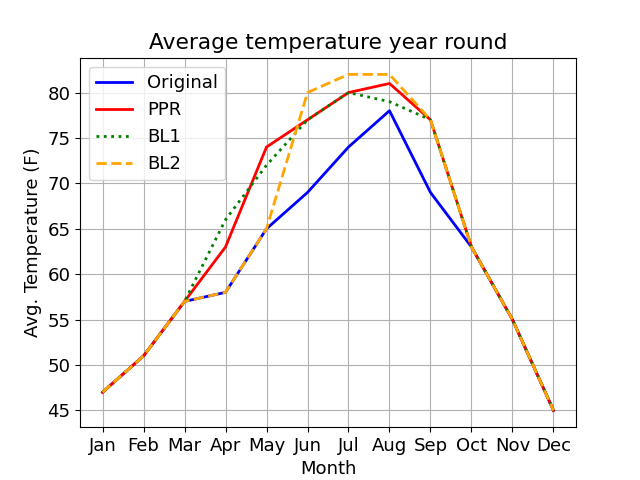}
\caption{}
\label{}
\end{subfigure}\hfill
\begin{subfigure}[t]{0.3\textwidth}
    \includegraphics[width=\textwidth]{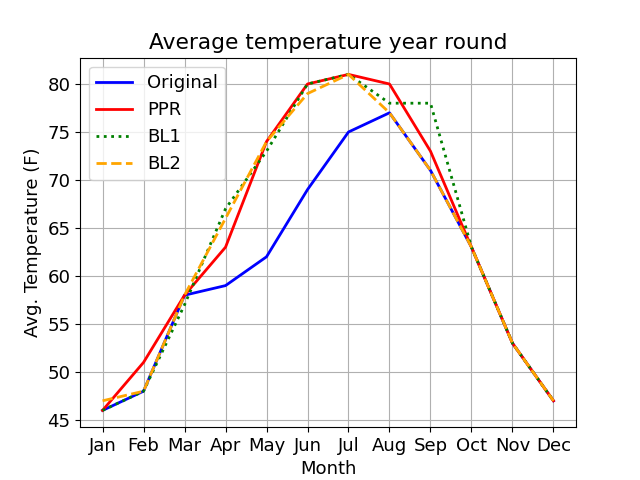}
\caption{}
\label{}
\end{subfigure}

\begin{subfigure}[t]{0.3\textwidth}
    \includegraphics[width=\linewidth]{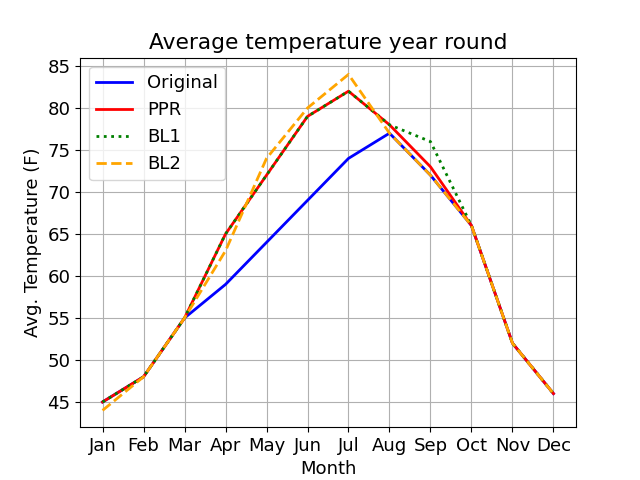}
\caption{}
\label{}
\end{subfigure}\hfill
\begin{subfigure}[t]{0.3\textwidth}
    \includegraphics[width=\linewidth]{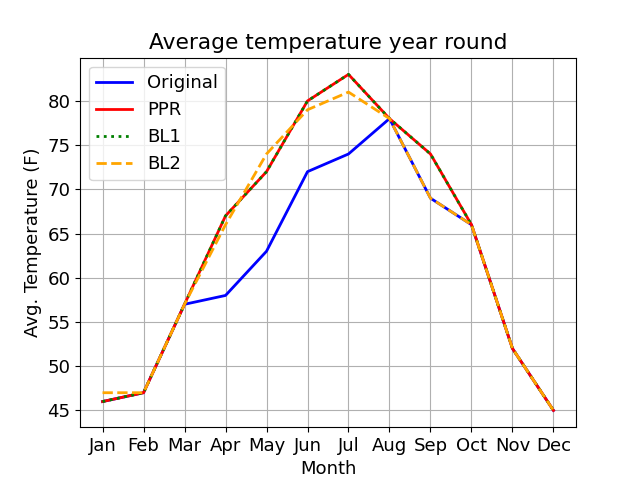}
\caption{}
\label{fig:figure14_8}
\end{subfigure}\hfill
\begin{subfigure}[t]{0.3\textwidth}
    \includegraphics[width=\linewidth]{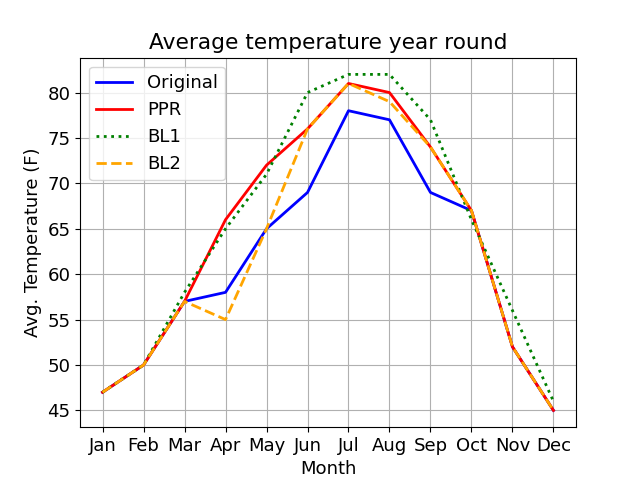}
% \caption{Original temperatures (from Rome) from the test set and the corresponding recourse paths from PPR, BL1, BL2. }
\label{fig:temperature2}
\end{subfigure}

\caption{Original temperatures (from Rome) from the test set and the corresponding recourse paths from PPR, BL1, BL2. }
\label{fig:temperature2}
\end{figure}

\begin{figure}[htbp]
\begin{subfigure}[t]{0.3\textwidth}
    \includegraphics[width=\linewidth]{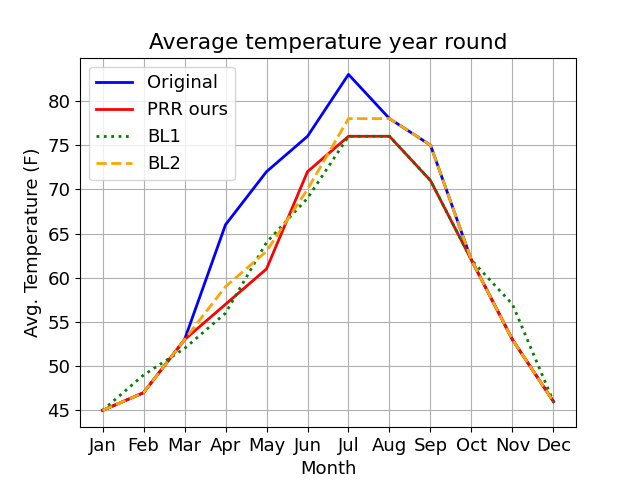}
\caption{}
\label{fig:figure14_1}
\end{subfigure}\hfill
\begin{subfigure}[t]{0.3\textwidth}
  \includegraphics[width=\linewidth]{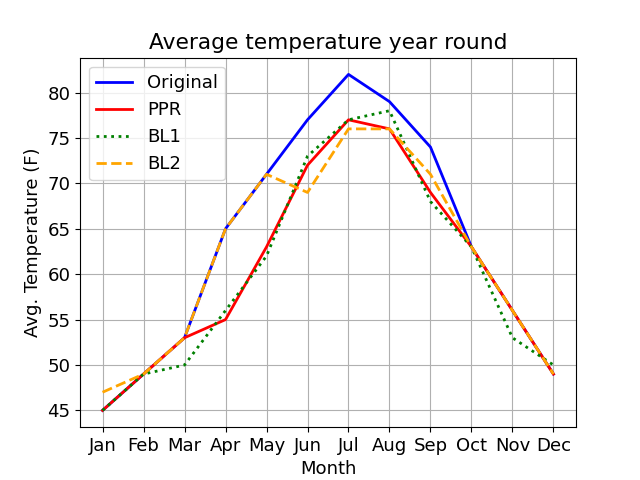}
\caption{}
\label{fig:figure14_2}
\end{subfigure}\hfill
\begin{subfigure}[t]{0.3\textwidth}
    \includegraphics[width=\linewidth]{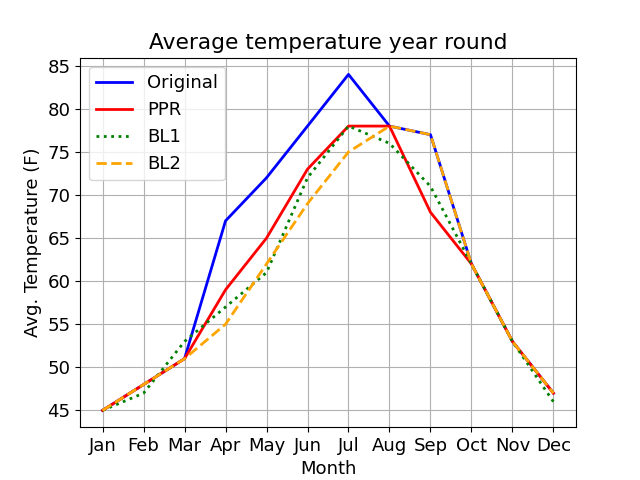}
\caption{}
\label{fig:figure14_3}
\end{subfigure}

\begin{subfigure}[t]{0.3\textwidth}
    \includegraphics[width=\linewidth]{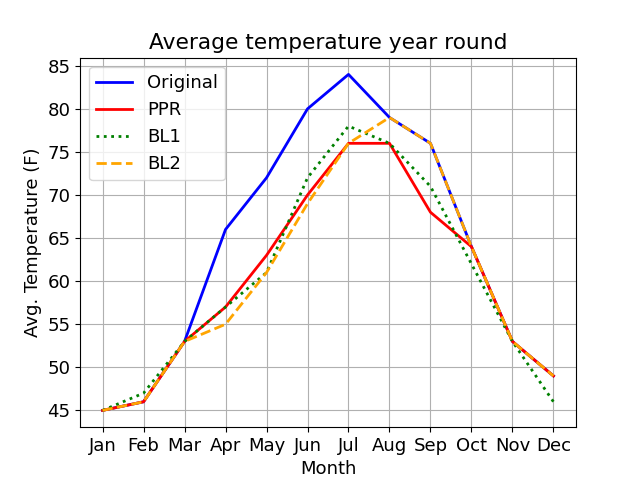}
\caption{}
\label{fig:figure14_4}
\end{subfigure}\hfill
\begin{subfigure}[t]{0.3\textwidth}
    \includegraphics[width=\linewidth]{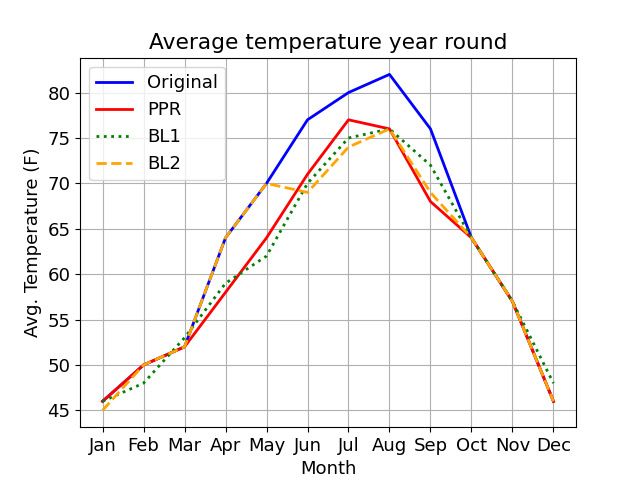}
\caption{}
\label{}
\end{subfigure}\hfill
\begin{subfigure}[t]{0.3\textwidth}
    \includegraphics[width=\textwidth]{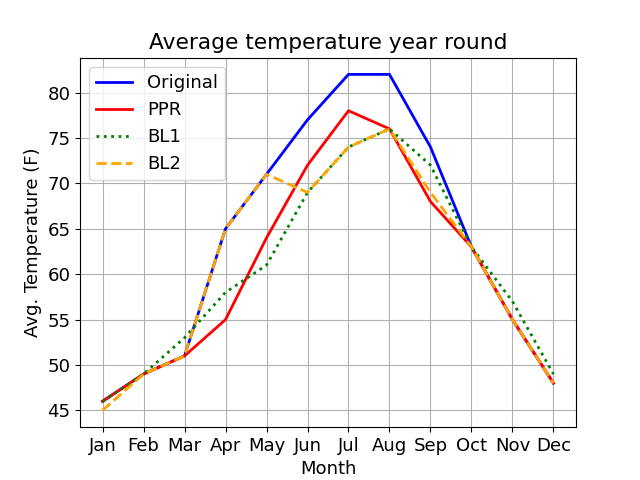}
\caption{}
\label{}
\end{subfigure}

\begin{subfigure}[t]{0.3\textwidth}
    \includegraphics[width=\linewidth]{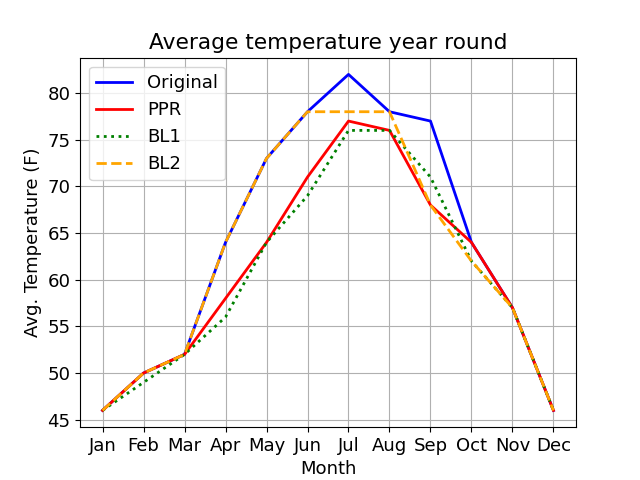}
\caption{}
\label{}
\end{subfigure}\hfill
\begin{subfigure}[t]{0.3\textwidth}
    \includegraphics[width=\linewidth]{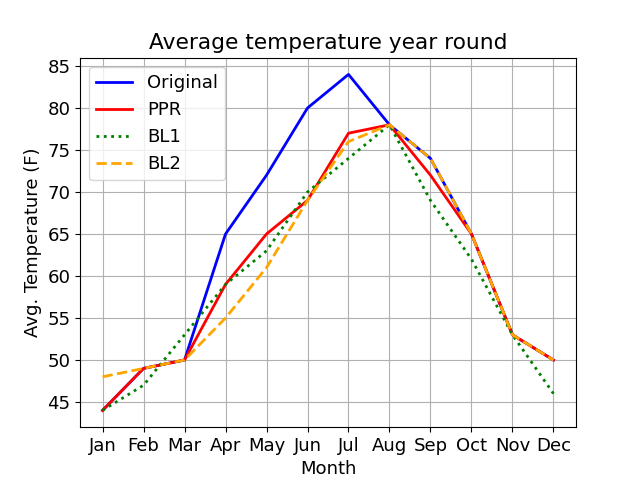}
\caption{}
\label{fig:figure14_8}
\end{subfigure}\hfill
\begin{subfigure}[t]{0.3\textwidth}
    \includegraphics[width=\linewidth]{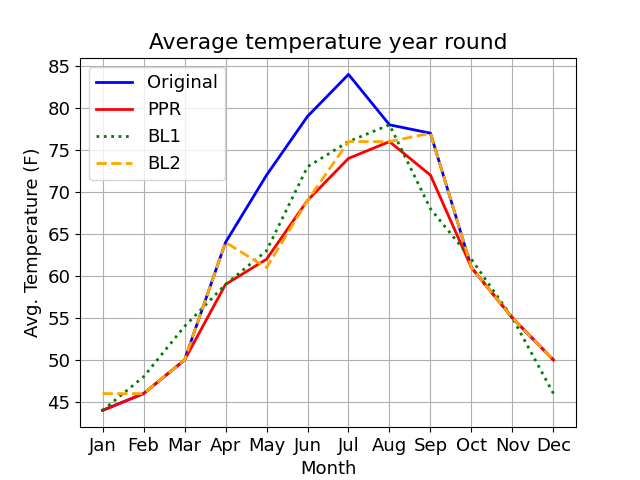}
\caption{}
\label{fig:figure14_9}
\end{subfigure}

\caption{Original temperatures (from Columbia) from the test set and the corresponding recourse paths from PPR, BL1, BL2. }
\label{fig:temperature3}
\end{figure}

\begin{figure}[h]
     \centering
     \begin{subfigure}[b]{0.3\textwidth}
         \centering
         \includegraphics[height = 3.5 cm, width=\textwidth]{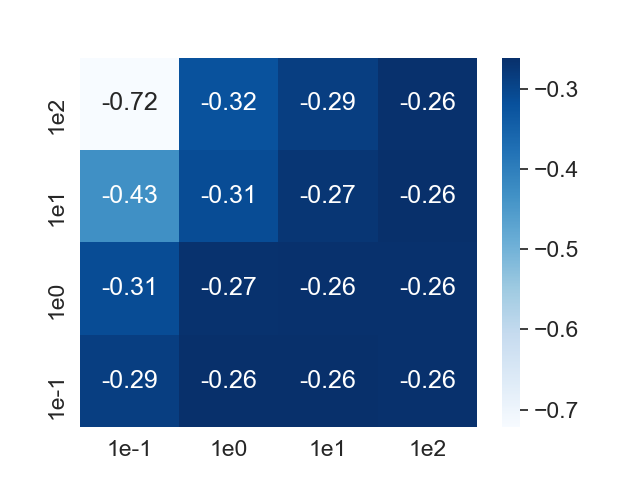}
         \caption{Personalization $s_\text{policy}$}
         % \label{fig:gridworld_original}
     \end{subfigure}
     \hfill
     \begin{subfigure}[b]{0.3\textwidth}
         \centering
         \includegraphics[height = 3.5 cm, width=\textwidth]{imgs/app1_heatmap_sim.png}
         \caption{Similarity $s_\text{path}$}
         % \label{fig:gridworld_experienced}
     \end{subfigure}
     \hfill
     \begin{subfigure}[b]{ 0.3\textwidth}
         \centering
         \includegraphics[height = 3.5 cm, width =\textwidth]{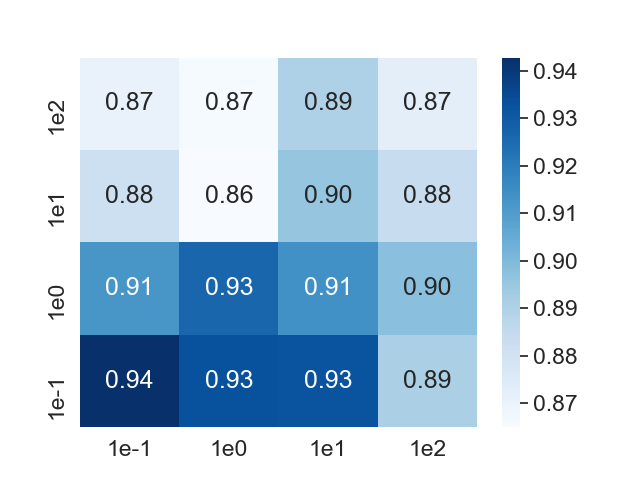}
         \caption{Goal score $s_\text{goal}$}
         % \label{fig:five over x}
     \end{subfigure}
        \caption{Personalization, similarity and goal score $s_\text{policy},s_\text{path},s_\text{goal}$ for the temperature experiment. The x-axis indicates different values of $\lambda_\text{path}$ and y-axis shows different values of $\lambda_\text{policy}$.
        }
        \label{fig:heatmap_temperature}
\end{figure}

\paragraph{Sensitivity Analysis} We also conduct sensitivity analysis with values for $\lambda_\text{policy},\lambda_\text{path}$.
Figure \ref{fig:heatmap_temperature} shows the heatmaps of the average score of $s_\text{policy},s_\text{path},s_\text{goal}$ based on different values of $\lambda_\text{policy},\lambda_\text{path}$.
Similar to the grid-world experiment, in general, the larger $\lambda_\text{policy}$ and $\lambda_\text{path}$, the higher $s_\text{policy}$ and $s_\text{path}$ are.
When both of these values are low, such as 0.01, the PPRs tend to have higher goal reward $s_\text{goal}$.

\paragraph{Baselines} For this application, a naive implementation of BL1 is still not applicable due to its high computational complexity.
For instance, if we allow the temperature ranges from 1 to 100, we will have $|\mathcal{A}| = 100$ and $|\mathcal{S}| = 100^{12}$, which are too large to fit in the memory for the dynamic programming method.
Therefore, we modified BL1 by only considering the available states and actions in the dataset.
On the other hand, BL2 aims to find the recourse sequence by first searching for the best candidate sequence with a different label from the original sequence.
Then, it replaces a subsequence in the original sequence with another subsequence extracted from this candidate.
Therefore, sometimes the generated recourse sequences contain some abnormal patterns.
For example, in figure \ref{fig:temperature3}(a), we can observe April's temperature is lower than March's in the recourse sequence generated by BL2.

\newpage
\newpage
\subsection{Text generation}

\paragraph{Dataset} In this experiment, we employ Transformer to train two language models: one on all 7 Harry Potter books by J.K. Rowling and another on the Bible corpus.
The link to the Harry Potter corpus can be found at \url{https://www.kaggle.com/code/balabaskar/harry-potter-text-analysis-starter-notebook} while the Bible text corpus is downloaded from \url{https://openbible.com/textfiles/asv.txt}.
We do text preprocessing before training: for the Harry Potter corpus, we merge all the books into one corpus.
For the Bible corpus, we remove all "Genesis" terms from the text files because of their redundancy.
We also convert all the text into lower-case before training.

\paragraph{Experiment set-up} The language models are trained with GPU NVIDIA GeForce GTX 1660 Ti. accelerator in three days.
Our transformer has an embedding dimension size of 200, the dimension of the feedforward network model is 200, 2 transformer encoding layers, and 2 multi-head attention.
We also set the dropout rate to 0.2.
We set $b=500,k=1,\gamma = 0.99,c_{e}=0.01,\text{decay} \: \epsilon=0.001,C=1$.
In this experiment, we modify values of $\lambda_\text{route},\lambda_\text{path}$ before generating the recourse text.

\paragraph{Examples of original texts and recoursed texts}
% Table \ref{tab:text-cf2} shows more examples of the recourse texts generated from different original texts, which flip the sentiment label from negative to positive.
In addition to Table \ref{tab:text-cf}, Table \ref{tab:text-cf3} shows examples of the recourse texts that flip the sentiment labels from positive to negative.

% \begin{table}[h]
%   \caption{Extra examples of text recourse with different writing styles that flip the sentiment labels from negative to positive. }
%   \label{tab:text-cf2}
%   \centering
%   \small
%   \begin{tabular}{p{2cm}p{4.9cm}p{4.9cm}}
%   % {|p{4cm}|p{5cm}|p{5cm}|}
%     \toprule

%     Original text     & $\pi_A$ trained from Harry Potter corpus    & $\pi_A$ trained from Bible corpus \\
%     \toprule
%     \multirow{5}{*}{She was sad.} & She was very pleased.  & She said my soul loveth.  \\
%     \cmidrule{2-3}
%      % & She was smiling happily  & $(1,1)$     \\
     
%      & She was smiling happily.  &   She called thee her blessed because she loved. \\
%      \cmidrule{2-3}
%      & She was so relieved.  &   She loved here, rachel, her children. \\
%      \toprule
%     \multirow{3}{2cm}{The cup was empty.}     & The cup of tea and whiskey, but he looked simply delighted. &       
%           The cup, he's been so happy     \\
%     \cmidrule{2-3}
%     & The cup which I found both great and small. &   The cup which is found, hidden part of heaven.  \\
%         \toprule
%     \multirow{4}{2.2cm}{The book on the table is boring.}      & The book is a very good one, said Hermione enthusiastically & The book of the law of the Lord, which he commanded. 
%     \\
%     \cmidrule{2-3}
%     & The book of monsters, which he was sure this would have been very good.      & The book of the chronicles of the kings of Judah and Jehoiada strengthen himself against Israel. \\
%     % & The book that had been very good in there      & $(1,5)$  \\
%     \bottomrule
%   \end{tabular}
% \end{table}

\begin{table}[h]
  \caption{More examples of text recourse with different writing styles that flip the sentiment from positive to negative. }\smallskip
  \label{tab:text-cf3}
  \centering
  \begin{tabular}{p{11cm}}
  % {|p{4cm}|p{5cm}|p{5cm}|}
    \toprule
    Original text and text sampled from $\mathcal{P}_A$ trained on Harry Potter and Bible corpus (with the corresponding $\lambda_\text{path}$ value)  \\   
    \toprule
        \textbf{Original text:} I am in love. \\
    \textbf{$\mathcal{P}_A$ trained from Harry Potter corpus:}
    \begin{itemize}
        \item $\lambda_\text{path} = 10$: I am sorry, it is my fault.
        \item $\lambda_\text{path} = 0.1$: "I didn't. Feel so stupid", said Ron. 
    \end{itemize}

        \textbf{$\mathcal{P}_A$ trained from Bible corpus:}
    \begin{itemize}
        \item $\lambda_\text{path} = 10:$ I am in distress.
        \item $\lambda_\text{path} = 0.1:$ I have seen my face and my sorrow is stirred.
    \end{itemize}\\
     \toprule
            \textbf{Original text:} The world is beautiful. \\
    \textbf{$\mathcal{P}_A$ trained from Harry Potter corpus:}
    \begin{itemize}
        \item $\lambda_\text{path} = 10$: The world is very nosy.
        \item $\lambda_\text{path} = 0.1$: The world and the cup is not strong. 
    \end{itemize}

        \textbf{$\mathcal{P}_A$ trained from Bible corpus:}
    \begin{itemize}
        \item $\lambda_\text{path} = 10$: The world is fallen.
        \item $\lambda_\text{path} = 0.1$: The world were crucified with him but saved us alive.
    \end{itemize}\\
        \toprule
            \textbf{Original text:} The magicians enjoy the magic. \\
    \textbf{$\mathcal{P}_A$ trained from Harry Potter corpus:}
    \begin{itemize}
        \item $\lambda_\text{path} = 10$: The magician had not committed this crime.
        \item $\lambda_\text{path} = 0.1$: The magic was expelled at Hogwarts. 
    \end{itemize}

        \textbf{$\mathcal{P}_A$ trained from Bible corpus:}
    \begin{itemize}
        \item $\lambda_\text{path} = 10$: The magicians of Egypt as an adversary and did as an evil man.
        \item $\lambda_\text{path} = 0.1$: The magicians of Egypt, but behold, they will bring their evil against their houses.
    \end{itemize}\\
    \bottomrule
  \end{tabular}
\end{table}

\paragraph{Baselines}
Remind that neither BL1 nor BL2 can be applied to this application. BL1 faces practical challenges because of its high computational complexity of $\mathcal{O}(n^2 mTk)$ ($n,m, T,k$ are the number of states, the number of actions, sequence length and the number of actions changed, respectively). For text data, both states and action space are extremely large, making BL1 computationally infeasible and creating a timeout error. BL2 is also not applicable due to the unavailability of a labeled text dataset, while PPR can directly work with a trained sentiment classifier.
\newpage

\section{Model Design Choices}
\subsection{Levenshtein distance formula}
The Levenshtein distance between two sequences 
$\tau_0 = <\mathbf{s}^{\tau_0}_1,\mathbf{s}^{\tau_0}_2,...,\mathbf{s}^{\tau_0}_{i}>$ and
$\tau_r = <\mathbf{s}^{\tau_r}_1,\mathbf{s}^{\tau_r}_2,...,\mathbf{s}^{\tau_r}_{j}>$ with length $i$ and $j$ respectively is defined as:
\[ 
d_{ij}(\tau_0,\tau_r)  = \left\{
\begin{array}{ll}
      d_{i-1,j-1} & \mathrm{if} \:\: {\mathbf{s}_i^{\tau_0}}
      
      = \mathbf{s}_j^{\tau_r} \\
       \mathrm{min} (
      d_{i-1,j} + w_{del}(\mathbf{s}_i^{\tau_0}), \\
       d_{i,j-1} + w_{ins}(\mathbf{s}^{\tau_r}_j), \\
      d_{i-1,j-1} + w_{sub}(\mathbf{s}_i^{\tau_0},\mathbf{s}^{\tau_r}_j) ) & \mathrm{if} \:\: \mathbf{s}_i^{\tau_0} \neq \mathbf{s}_j^{\tau_r}
\end{array} 
\right. 
\]
Here $w_{del}(\cdot),w_{ins}(\cdot)$ and $w_{sub}(\cdot, \cdot)$ are the functions returning the weighting scores of deletion, insertion, and substitution operation, respectively.
In this paper, we set all those values to 1.

\subsection{Personalization reward design}
In this section, we explain the design of Equation (4) of the $h(.)$ function in the main paper.
As a reminder, we design a link function $h(\cdot)$ that needs to satisfy the conditions as follows: 
\begin{enumerate}
\item  $h(p)$ increases monotonically with probability $p$.
\item $h(p)>0$ when $p > \frac{1}{|\mathcal{A}|}$, $h(p) = 0 $ when $p = \frac{1}{|\mathcal{A}|}$, and $h(p)<0$ when $p < \frac{1}{|\mathcal{A}|}$.
\item  When $p$ is close to 1, $h(p)$ is a large positive number and when $p$ is close to 0, $h(p)$ is a large negative number.
\end{enumerate}

In addition, we would like to penalize $h(.)$ when it reaches a too high or too 
 low value by adding $\log()$ factor.
Then $h(.)$ has the form:
\begin{equation}
    h(p) = \log(\frac{f(p)}{g(p)})
\end{equation}

$h(p)$ must increase monotonically with probability $p$. 
For simplicity, we assume $f(p),g(p)$ are linear functions, so we can write $h(p)$ 
 as:

 \begin{equation}
       h(p) = \log(\frac{A_1 p + B_1}{A_2 p + B_2})         
 \end{equation}

% p > $\frac{1}/{|\mathcal{A}|}$
First, assuming $\frac{1}{|\mathcal{A}|} < p < 1$, when $p$ is close to 1, we want $h(p)$ to have high positive number. 
This can be achieved by set $g(p) = A_2 p + B_2 = 0$ when $p = 1$, which is equivalent to $A_2 = -B_2$. 
Plus the condition $\frac{1}{|\mathcal{A}|} < p < 1$, then we can choose $B_2 = 1$ and $A_2 = -1$.
The final form of $g(p)$ is:
\begin{equation}
    g(p) =  1 - p
\end{equation}

Furthermore, we want $h(p)$ to be 0 when $p = \frac{1}{|\mathcal{A}|}$.
From (8), this is equivalent to $\frac{f(p)}{g(p)} = 1$ or $f(p) = g(p)$ when $p = 1/{|\mathcal{A}|}$. 
Replace with (10), we have:
\begin{equation}
    A_1 p + B_1 = 1 - p \: \: \text{when} \: \:  p = 1/|\mathcal{A}|
    \Longleftrightarrow
    \frac{A_1}{|\mathcal{A}|} + B_1 = 1 - \frac{1}{|\mathcal{A}|} 
    \Longleftrightarrow
    B_1 = 1 - \frac{1 + A_1}{|\mathcal{A}|}
\end{equation}

If we choose $A_1 = 1$ then $B_1 = 1 - \frac{2}{|\mathcal{A}|}$.

So when $\frac{1}{|\mathcal{A}|} < p < 1$, the final form of $h(.)$ is:
\begin{equation}
    h(p) = \log(\frac{p - \frac{2}{|\mathcal{A}|} + 1}{1 - p})
\end{equation}

Next, assuming $0 < p < \frac{1}{|\mathcal{A}|}$.
With similar reasoning, we want $h(p)$ to be a very low negative value when $p = 0$.
This is a natural property of $\log(.)$ function, so $h(.)$ can have the form:
\begin{equation}
    h(p) = \log(A \: p)
\end{equation}
We want $h(p) = 0$ when $p = 1/|\mathcal{A}|$, this is equivalent to $A p = 1$ when $p = 1/|\mathcal{A}|$, or $\frac{A}{|\mathcal{A}|} = 1 \Longleftrightarrow A = |\mathcal{A}|$.
Then, $h(p)$ has the form:
\begin{equation}
    h(p) = \log(|\mathcal{A}| p)
\end{equation}

So, to sum up, the final form of $h(p)$ defined on domain [0,1] is:
 \begin{equation}
            h(p)=
\begin{cases}
\log(\frac{p - \frac{2}{|\mathcal{A}|} + 1}
    {1-p}) 
 & \text{if $p >= \frac{1}{\mathcal{|A|}}$,} \\
\log(|\mathcal{A}| \cdot p) & 
\text{otherwise}.
\end{cases}
    \end{equation}

which is Equation (4) in the main paper.
Figure 3 illustrates the graph of $h()$ function.
\begin{figure}[h]
\includegraphics[width=4cm, height = 5cm]{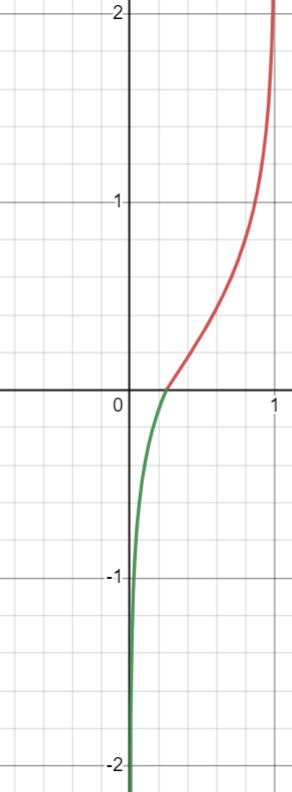}
\centering
\caption{Visualization of $h()$ function}
\end{figure}

\end{document}